%% file: colm2026_conference.tex
\documentclass{article} 
\usepackage[final]{colm2026_conference}

\usepackage{microtype}
\usepackage{hyperref}
\usepackage{url}
\usepackage{booktabs}

\usepackage{graphicx}

\usepackage{booktabs}

\usepackage{amsmath}
\usepackage{algorithm}
\usepackage{algpseudocode}
\usepackage{amsfonts}
\usepackage{multirow}
\usepackage{subcaption}
\usepackage{xcolor}
\usepackage{cleveref}

\usepackage{lineno}

\definecolor{darkblue}{rgb}{0, 0, 0.5}
\hypersetup{colorlinks=true, citecolor=darkblue, linkcolor=darkblue, urlcolor=darkblue}

\title{Multi-Granular Node Pruning for Causal Circuit Discovery}

\author{
Muhammad Umair Haider\textsuperscript{1}, \quad
Hammad Rizwan\textsuperscript{2}, \quad
Hassan Sajjad\textsuperscript{2}, \quad
A.B.~Siddique\textsuperscript{1} \\[4pt]
\textsuperscript{1}Department of Computer Science, University of Kentucky, USA \\
\textsuperscript{2}Department of Computer Science, Dalhousie University, Canada \\[2pt]
Correspondence: \texttt{muhammadumairhaider@uky.edu}
}

\begin{document}

\ifcolmsubmission
\linenumbers
\fi

\maketitle

\begin{abstract}
Causal circuit discovery aims to identify minimal subnetworks that causally drive specific behaviors in large language models (LLMs). 
Existing approaches focus on edge pruning or unstructured weight pruning. 
These methods are computationally expensive and typically operate on coarse-grained components, such as attention heads or MLP blocks, thereby missing finer-grained structure. 
We propose a multi-granular node-level pruning framework \footnote{Code: \url{https://github.com/MuhammadUmairHaider/Node-Pruning-for-Circuit-Discovery}} for circuit discovery that addresses both scalability and granularity limitations. 
Our approach introduces learnable masks across multiple levels of granularity, from entire blocks to individual neurons, within a unified optimization objective. 
Granularity-specific sparsity penalties guide the pruning process, enabling multi-level compression in a single optimization run.
Empirically, our approach identifies more compact circuits than prior methods, pruning $33.34\%$ more MLPs, and $59.8\%$ more neurons in the least favorable setting, with substantially larger gains overall.
We further show that many neurons retained by coarse-grained methods are, in fact, unnecessary and can be removed with negligible impact on task performance.
Our method is also substantially more memory efficient, requiring at least \(3\times\) less memory by avoiding the need to store intermediate activations during pruning.
\end{abstract}

\section{Introduction}
\label{sec:intro}

Large language models (LLMs) have demonstrated remarkable capabilities across a wide range of tasks, from question answering to code generation~\citep{brown2020language,chowdhery2023palm,achiam2023gpt}. 
However, their growing use in high-stakes applications has intensified concerns about interpretability and reliability~\citep{rudin2019stop,bommasani2021opportunities,weidinger2021ethical}. 
Because the internal mechanisms underlying their behavior remain poorly understood, it is difficult to trust, debug, and analyze these models~\citep{lipton2018mythos,rudin2019stop}.
Circuit discovery~\citep{olah2020zoom,cammarata2020thread} has emerged as a promising approach for addressing this challenge by isolating minimal subnetworks, or circuits, that causally drive specific behaviors within a larger model.

\begin{figure*}[t!]
\includegraphics[width=\textwidth]{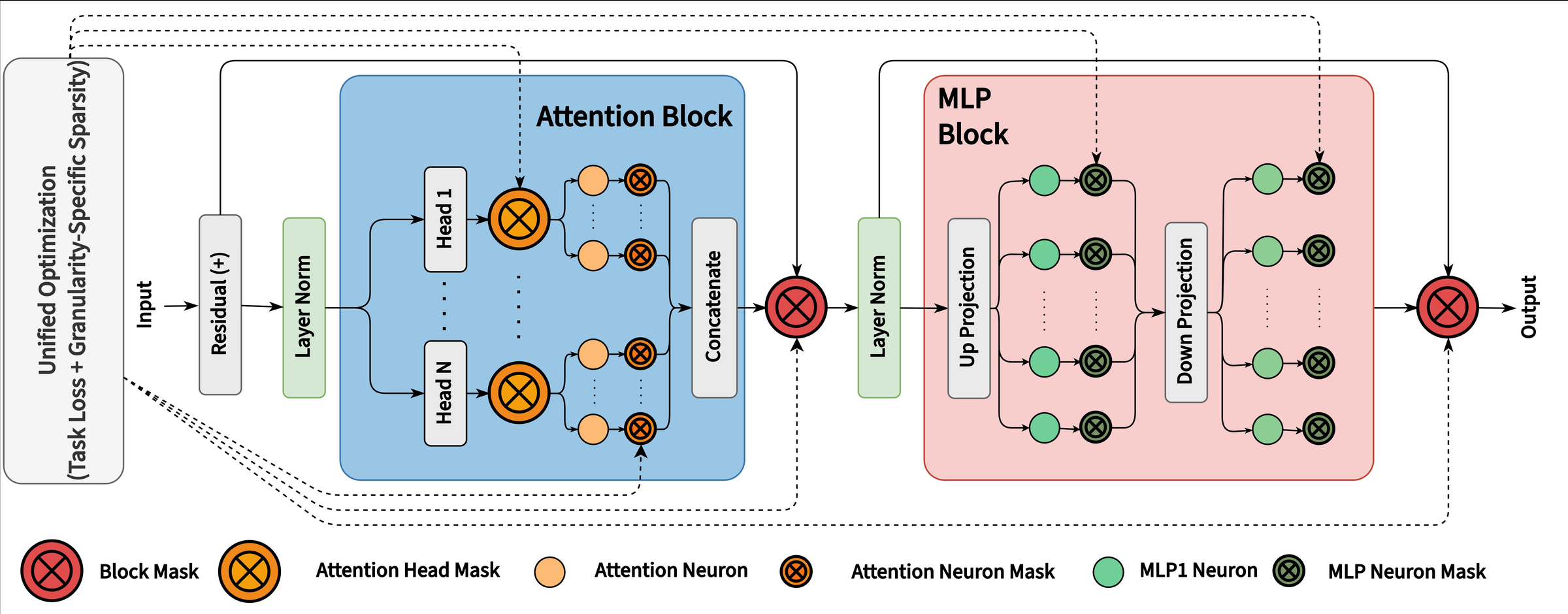} 
\vspace{-14pt}
\caption{Overview of the multi-granularity node pruning framework. Learnable masks are applied hierarchically across a transformer layer at block level (\textcolor{red}{red}), attention head level (\textcolor{orange}{orange}), and neuron level (\textcolor{green}{green}), spanning both the Attention Block and MLP Block (Up/Down projections). A unified optimization objective combines task-specific loss with a sparsity penalty. A two-stream configuration interpolates between clean states $h^{\text{clean}}$ and corrupted states $h^{\text{corrupted}}$ via a parameterized mask $m$.}
\label{fig1}
\vspace{-8pt}
\end{figure*}

Current circuit discovery methods predominantly rely on edge pruning, which removes connections between components to isolate circuits responsible for specific behaviors~\citep{wang2022interpretability,conmy2023towards,bhaskar2024finding}.
While these methods have advanced our understanding of model behavior, they suffer from two key limitations. 
First, edge pruning is computationally expensive: it requires storing \textit{disentangled} residual-stream representations and intermediate activations during pruning~\citep{bhaskar2024finding}, and often involves many forward passes through the model~\citep{conmy2023towards}.
Moreover, because models contain exponentially more edges than nodes, edge-level pruning becomes prohibitively expensive as models scale.
Second, these methods operate at a coarse granularity, treating entire attention heads or MLP blocks as atomic units.
This restricts pruning to the component level and overlooks the possibility that specific neurons (nodes) within these components can be responsible for the target behavior. 
As a result, recovered circuits can miss important fine-grained structures.
Recent work~\citep{yu2025sheaf} combines edge pruning with unstructured weight pruning by jointly learning masks over computation graph connections and individual weights. 
However, the induced sparsity remains largely unstructured and typically yields limited node-level pruning. 

In this work, we propose a multi-granular node-pruning framework for causal circuit discovery that addresses both the scalability and granularity limitations of prior methods.
Our framework learns masks over nodes at multiple levels of granularity, ranging from entire transformer blocks to attention heads, MLP blocks, and individual neurons, within a unified optimization objective, as illustrated in Figure~\ref{fig1}.
These masks are optimized in a two-stream causal pruning setup, where each node is softly interpolated between its clean and corrupted activation, and the resulting circuit is trained for faithfulness, task performance, and sparsity.


Across three standard circuit-discovery tasks, Indirect Object Identification (IOI), Gendered Pronouns (GP), and Greater Than (GT), on GPT-2, our approach consistently discovers substantially more compact circuits than prior approaches while preserving strong task fidelity. 
In the least favorable setting, 33.34\% more MLP blocks, and 59.8\% more neurons than prior methods, with larger gains in other settings. Unlike existing baselines, which achieve only limited MLP- and neuron-level compression, our framework identifies fine-grained circuit structures beyond coarse components. 
Our method is also substantially more compute-efficient, using \(3\times\) to \(11\times\) less GPU memory than prior methods and scaling circuit discovery to Llama 3.1-8B on a single 40GB GPU.

We analyze the discovered circuits and report several key findings. The structure of the discovered circuits depends on the task under consideration. For instance, most MLP blocks are essential for the IOI task, whereas for the GT task, most MLPs can be removed without significant performance degradation. Furthermore, our analysis indicates that different component granularities exhibit varying degrees of susceptibility to compression. For instance, the outputs of MLP1 (Keys) can be sparsified to a greater extent than those of MLP2 (Values)~\citep{geva2020transformer}.

The key contributions of this work are as follows:

\begin{itemize}
\item We introduce a multi-granular node-pruning framework for causal circuit discovery that learns masks at multiple levels of granularity, ranging from transformer blocks to individual neurons, within a single unified optimization objective.

\item We show that our framework identifies substantially more compact circuits than prior approaches, while also reducing GPU memory usage and scaling to larger models such as Llama 3.1-8B on a single 40GB GPU.

\item We find that the identified circuits exhibit task-dependent structure, showing that different behaviors are encoded by distinct patterns of attention, MLP, and neurons.


\end{itemize}

\section{Related Work}
\textbf{Manual Circuit Discovery.} Early mechanistic interpretability work manually traced through model computations to identify circuits. \citet{wang2022interpretability} discovered the indirect object identification (IOI) circuit in GPT-2, revealing how models track syntactic dependencies. \citet{olsson2022context} identified induction heads as a fundamental circuit pattern for in-context learning. While groundbreaking, these manual approaches require extensive human effort and domain expertise, limiting their applicability to larger models and diverse tasks.

\textbf{Automated Circuit Discovery.} To scale circuit analysis, prior work has developed automated methods based on iterative edge ablation.  Automated Circuit Discovery (ACDC) by \citet{conmy2023towards} formulates circuit finding as iteratively removing edges between components while maintaining task performance. The method starts with a full computational graph and greedily removes the least important edges, requiring O(n²) forward passes for n edges. \citet{syed-etal-2024-attribution} extend this with Edge Attribution Patching (EAP), using gradient-based importance scores to guide edge removal, requiring a single gradient pass to calculate edge importance. While these methods successfully automate circuit discovery, they suffer from computational intractability for large models, and results are limited to connections between coarse-grained components like attention heads and MLP blocks.


\textbf{Learnable Masking Approaches.} Recent work has explored learnable masks for circuit discovery. \citet{bhaskar2024finding}'s Edge Pruning (EP) introduces differentiable masking (CoFi
masking~\citep{xia2022structured}) for edges, to learn connection importance. They also opt to perform coarse-grained node pruning primarily for training stability, but as shown in \S~\ref{sec:results}, it does not yield compact circuits. Disco-GP~\citep{yu2025sheaf} learns masks over computation graph connections along with unstructured parameter sparsity. However, these approaches still primarily operate on edges between coarse components, or unstructured weights removal, unstructured weight sparsity does not reliably induce node-level sparsity, since a neuron remains active unless all its incoming connections are pruned. Our work differs fundamentally by applying learnable masks to nodes at multiple granularities, from entire blocks down to individual neurons.


\textbf{Traditional Pruning.} Traditional pruning methods~\citep{han2015learning, lecun1989optimal, haider2021comprehensive} target model compression for efficient deployment, removing weights or modules with minimal impact on overall performance. These methods yield hardware-aligned sparsity patterns and permanently remove selected components. Circuit discovery differs fundamentally: it seeks a minimal subnetwork responsible for a \emph{particular behavior}, prioritizing interpretability over deployment efficiency, and uses adversarial patching rather than zero ablation to causally validate the circuit's role.

\section{Preliminaries}
 


\textbf{Circuit Discovery Formulation.} Given a transformer model $M$ with parameters $\theta$ and a specific task $T$, circuit discovery seeks to find a minimal subgraph $G^* = (N^*, E^*)$ of the model's computational graph $G = (N, E)$ such that the subgraph preserves task performance. The nodes $N$ represent components at various granularities, from entire layers and attention heads down to individual neurons and edge set $E$ represents the computational dependencies between them.


The circuit discovery objective is then formulated as:
\begin{equation}
G^* = \arg\min_{G' \subseteq G} |G'| 
\quad \text{s.t.} \quad 
\mathcal{L}_T(G') \leq \mathcal{L}_T(G) + \epsilon,
\end{equation}
where $|G'|$ measures the size of the subgraph (e.g., number of nodes or edges), 
and $\epsilon$ denotes an allowable performance tolerance.



\textbf{Prior Formulations and Limitations.} Edge pruning approaches such as ACDC~\citep{conmy2023towards}, EP~\citep{bhaskar2024finding}, and DiscoGP~\citep{yu2025sheaf} formulate circuit discovery as an edge-pruning process. The common formulation underlying these approaches is to start from the full graph $G$, and remove all edges whose removal preserves the loss within a tolerance $\epsilon$, yielding the pruned graph:
\begin{equation}
G^* = G \setminus E_{\mathrm{rem}}^*, 
\qquad
E_{\mathrm{rem}}^* = \arg\max_{E' \subseteq E} |E'|
\ \text{s.t.} \
\mathcal{L}_T\!\left(G \setminus E'\right) \leq \mathcal{L}_T(G) + \epsilon.
\end{equation}
where $|E'|$ denotes the number of removed edges.
To reduce computational overhead, these methods operate at a \emph{component-level granularity}, 
where the graph nodes $N$ consist of attention heads and MLP blocks, and the edge set $E \subseteq N \times N$ encodes all possible connections between them. 

As every node can potentially connect to every other downstream node, the number of edges grows quadratically with the number of nodes, i.e., $|E| = O(|N|^2)$.  Therefore, exploring finer granularities, in terms of edges, would require searching over a vastly larger space. This quadratic scaling renders such approaches computationally intractable as the model size scales.

\section{Multi-Granular Node Pruning}

We present a unified, node-level pruning framework that discovers minimal circuits across multiple granularities in a single optimization step. The framework employs learnable masks applied at different levels of granularity and uses a two-stream forward pass that interpolates between clean and corrupted activations to estimate the contribution of each component to task performance.

\subsection{Granularity Design}

Recent results in mechanistic interpretability suggest that meaningful computation in transformers is often organized below the level of standard components (Attention Heads or MLPs). 
Individual neurons can implement specific features, attention heads frequently mix multiple unrelated functions, and many behaviors are mediated by sparse subsets of neurons rather than entire layers or blocks~\citep{cunningham2023sparse, dar2022analyzing, wang2022interpretability, haider2025neurons}. These observations motivate circuit discovery methods that go beyond edges between attention heads and MLP blocks. Since transformers have a naturally hierarchical organization, they support discovering circuits at multiple granularities, from coarse components down to fine-grained neuron-level substructures.

In this work, we implement a hierarchical multi-granularity circuit discovery framework spanning distinct levels of granularity:
\[\mathcal{V}\!=\!\{V^{(b)}_{\text{attn}},V^{(b)}_{\text{mlp}},V^{(h)}_{\text{attn}},V^{(n)}_{\text{attn}},V^{(n)}_{\text{mlp1}},V^{(n)}_{\text{mlp2}}\}
\]
where \(V^{(b)}_{\text{attn}}\) and \(V^{(b)}_{\text{mlp}}\) denote attention and MLP blocks, \(V^{(h)}_{\text{attn}}\) denotes attention heads, \(V^{(n)}_{\text{attn}}\) denotes attention neurons, and \(V^{(n)}_{\text{mlp1}}, V^{(n)}_{\text{mlp2}}\) denote neurons in the first and second linear layers of the MLP, respectively.


\subsection{Two-Stream Causal Pruning Setup}

In contrast to traditional pruning methods that simply zero out components, our approach identifies circuit elements by evaluating their importance through targeted corruption in line with \citet{conmy2023towards,chan2022causal,bhaskar2024finding}. 
Zero and mean ablation introduce distribution shifts, resulting in pruned circuits that can reflect compensatory behavior rather than the mechanism used under normal operation. The pruning process might simply learn a new solution, akin to the Strong Lottery Ticket Hypothesis~\citep{sreenivasan2022finding,ramanujan2020s}, rather than uncovering the true underlying mechanism. In contrast, corrupted patching is actively adversarial; by swapping task-critical information (e.g., inverting gender pronouns), it attempts to deceive the model. This ensures that the discovered circuit is composed of components that explicitly discern between the correct and incorrect behavior~\citep{chan2022causal,hanna2023does}.

\textbf{Clean Input Stream.} 
  Feed the model an input designed to elicit a specific, notable behavior. For example, for the prompt ‘The war started in 1950 and ended in 19\_\_.’, the model should place higher probability mass on numbers greater than 50 than on numbers less than 50.

\textbf{Corrupted Input Stream.} 
    Passes a minimally perturbed version of the same input to the model, using the task-dependent corruption function $C$, through the model. The corruption preserves input length but alters specific tokens to elicit different model predictions with distinct correct answers. Crucially, both inputs correspond to the same task type, enabling the identification of shared circuitry by observing how the same components are differentially activated~\citep{vig2020investigating, meng2022locating, wang2022interpretability}. For instance, \textit{"The war started in 1901 and ended in 19\_\_."} is a corrupted perturbation of \textit{"The war started in 1950 and ended in 19\_\_."}, In the corrupted instance, the model is no longer required to assign a higher probability to the numbers greater than 50.


\textbf{Forward Pass Formulation.}
For each node \( n_i \in N \), we define a parameterized mask \( m_i \in [0, 1] \) that is optimized to guide the model toward producing the correct output by selectively interpolating clean-input activations with activations from the corrupted input. The output of each node is computed as:
\vspace{-6pt}
\begin{equation}
\begin{split}
\label{eq:forward_pass}
&h_i^{node} = I(h_i^{\text{clean}}, h_i^{\text{corrupted}}, m_i) \\
&I(.) 
  = m \cdot h^{\text{clean}} + (1 - m) \cdot h^{\text{corrupted}}
\end{split}
\end{equation}

where \( h_i^{\text{clean}} \) and \( h_i^{\text{corrupted}} \) denote the clean and corrupted hidden activations of node \( n_i \), respectively. When \( m_i = 1 \), the node passes the clean activation; when \( m_i = 0 \), it passes the corrupted activation. Intermediate values \( 0 < m_i < 1 \) represent soft interpolation during training. 
The set of all masks \( M \) is regularized to encourage sparsity, ensuring that only the nodes most relevant to the task remain active. 

This formulation identifies circuits by finding components whose corruption significantly impacts task performance. The key hypothesis is that important components will show a large shift in performance between clean and corrupted activations, whereas unimportant components will show minimal differences.

\textbf{Mask Parameterization.} To enable gradient-based optimization while inducing approximately binary behavior, we parameterize the masks using the \textit{Hard Concrete} distribution~\citep{louizos2017learning}. 
The sampling procedure is defined as:
\vspace{-6pt}
\begin{equation}
\begin{split}
\label{eq:hard_concrete_mask}
&m_{i} = H(\alpha_{i},\beta) \\
&H(\alpha_{i},\beta) = \min(1, \max(0, s_{i} (\zeta -\gamma)  + \gamma))\\
&s_{i} = \sigma\left(\frac{1}{\beta}(\log u - \log(1-u) + \log \alpha_{i})\right) \\
&u_i  \sim \text{Uniform}(0, 1).
\end{split}
\end{equation}

Here, \(\alpha_i\) is a trainable parameter for the \(i^{\text{th}}\) gate, \(\beta\) controls the smoothness, and \(\zeta>1\) and \(\gamma<0\) define the support range. We use the standard values \((\zeta=1.1,\ \gamma=-0.1,\ \beta=0.66)\) from \citet{louizos2017learning}. Each layer contains multiple such gates applied to different components (e.g., blocks, heads, or neurons).

\subsection{Training Objective}

Our objective encourages finding minimal circuits that maintain task performance:
\vspace{-8pt}
\begin{equation}
\begin{split}
&\mathcal{L} = (1-\lambda)(\mathcal{L}_{\text{faith}}(h^{\text{base}},h^{\text{circuit}})
+ \mathcal{L}_{\text{task}}(h_{gt}^{\text{circuit}},h_{distr}^{\text{circuit}})) + \lambda \sum_{i=1}^{N} \mathcal{L}_0(m_i)  \label{eq:loss}\\
&\mathcal{L}_{\text{faith}} =
\mathrm{KL}\!\left(
\mathrm{softmax}(h^{\text{base}}) \,\big\|\, 
\mathrm{softmax}(h^{\text{circuit}})
\right) \\
&\mathcal{L}_{\text{task}} =
\max\!\left(0,\; m-\left(h_{gt}^{\text{circuit}}-h_{distr}^{\text{circuit}}\right)\right) \\
&\mathcal{L}_0(m_i) =
\sigma\!\left(\log \alpha_{i} - \beta \log \frac{-\gamma}{\zeta}\right) \\
\end{split}
\end{equation}

where $h^{\text{base}}$ denotes the base model representation and
$h^{\text{circuit}}$ denotes the representation produced by the current circuit.
$h_{gt}^{\text{circuit}}$ and $h_{distr}^{\text{circuit}}$ denote the current circuit logits
for the ground-truth and distractor options, respectively.
$\mathcal{L}_{\text{faith}}$ is the faithfulness loss (a KL divergence) that encourages
the circuit's output distribution to match the base model's distribution,
$\mathcal{L}_{\text{task}}$ is a margin-based ranking loss with margin $m$,
and $\lambda$ controls the strength of the sparsity regularizer
$\sum_{i=1}^{N}\mathcal{L}_0(m_i)$. The training pseudocode is provided in Algorithm~\ref{alg:node_pruning}

\subsection{Circuit Extraction}

After training, we binarize all masks applying the following constraint:
\vspace{-2pt}
\begin{equation}
\label{eq:hard_concrete_final}
m_i^{\text{final}} = \mathbf{1}\!\left(\log \alpha_i > \beta \log\!\left(\tfrac{-\gamma}{\zeta}\right)\right),
\end{equation}

Furthermore, we enforce hierarchical consistency among masks:\textit{ if a parent mask is deactivated, all its associated child masks are also set to zero.} For instance, when an MLP block is pruned (mask set to zero), the neuron masks within that block are likewise disabled.

\section{Experimental Setup}

\subsection{Task and Task Specific Metrics}
\label{sec:tasks}
We evaluate our method on three standard circuit discovery tasks, which use an adversarial formulation, that test different computational capabilities within language models. 

\textbf{Greater-Than (GT).} This task probes the numerical reasoning capabilities of language models. For instance, given \textit{``The war lasted from the year 1743 to the year 17\_\_''}, the model should prefer completions 44--99 over 00--42. Using the dataset of \citep{hanna2023does}, we measure this behavior by $\mathrm{GT} = \mathbb{E}_{x \in \mathcal{D}_{\text{test}}} \!\left[ P(y > y_{\text{start}} + 10 \mid x) - P(y < y_{\text{start}} - 10 \mid x) \right]$, where $y_{\text{start}}$ is the year in the prompt. Here, the two terms denote the total probability mass assigned to completions more than 10 years after and before $y_{\text{start}}$, respectively; positive values indicate a stronger preference for valid future years. We use a 10 years window, so that the circuit does not exploit a shortcut of always predicting a higher number like 99.


\textbf{Indirect Object Identification (IOI).} This task probes whether a language model can track entities and identify the correct recipient in context. For example, given \textit{``Friends Juana and Kristi found a mango at the bar. Kristi gave it to''}, the model should prefer \textit{Juana} over the repeated subject \textit{Kristi}. We measure this with $\mathrm{IOI\text{-}Score} = \mathbb{E}_{x \in \mathcal{D}_{\text{test}}} \!\left[ \mathrm{logit}_{\mathrm{IO}}(x) - \mathrm{logit}_{\mathrm{S}}(x) \right]$, where $\mathrm{logit}_{\mathrm{IO}}(x)$ and $\mathrm{logit}_{\mathrm{S}}(x)$ denote the logits of the correct indirect object and incorrect repeated subject, respectively. Positive values indicate correct disambiguation.


\textbf{Gendered Pronouns (GP).} This task probes learned associations between names and pronouns. For example, given \textit{``So Evan is a really great friend, isn't ''}, the model should assign higher probability to \textit{he} than to \textit{she}. We measure this with $\mathrm{GP\text{-}Score} = \mathbb{E}_{x \in \mathcal{D}_{\text{test}}} \!\left[ \mathrm{logit}_{\mathrm{consistent}}(x) - \mathrm{logit}_{\mathrm{inconsistent}}(x) \right]$, where the two terms denote the logits of the gender-consistent and gender-inconsistent pronouns, respectively. Positive values indicate stronger gender-consistent associations.


\subsection{General Metrics}

\textbf{Task Accuracy:} Based on task-specific metrics previously discussed we also compute task \textit{accuracy} by binarizing scores: positive scores map to 1, and non-positive scores to 0.

\textbf{Distribution-level faithfulness}, we compute the Kullback-Leibler (KL) divergence between the full model's output logit distribution ($P_m$) and the pruned circuit's logit distribution ($P_c$):
\begin{equation}
\mathcal{D}_{\text{KL}}(P_{\text{m}} || P_{\text{c}}) = \sum_{v \in \mathcal{V}} P_{\text{m}}(v) \log \frac{P_{\text{m}}(v)}{P_{\text{c}}(v)}
\end{equation}
where $\mathcal{V}$ is the vocabulary and $P$ represents the probability distribution over next tokens. This metric quantifies how closely the circuit reproduces the full model’s predictive distribution~\citep{bhaskar2024finding,conmy2023towards}.

\textbf{Circuit Size Metrics.} We report circuit size across multiple granularities using the following size metrics:
\textit{(1) Sparsity by level} -- percentage pruned at each granularity (layers, heads, neurons).
\textit{(2) Edge count comparison} -- we convert our node-based circuits into equivalent edge counts for fair comparison with edge-pruning baselines.

As edge-pruning baselines do not natively report node-level results, and our method does not natively report edge-level results, we employ strict proxy checks to facilitate a fair comparison across granularities.
For \textit{node sparsity of edge-pruning methods}, we consider a node pruned only if it is an ``island'' with no incoming or outgoing edges. For attention components, we count a node as part of the circuit only when its value stream is retained. We then perform a reverse Breadth-First Search (BFS) to remove all isolated islands, reporting only the surviving nodes.
For \textit{edge sparsity of our method}, we report a conservative estimate by assuming all remaining nodes are fully connected.

Together, these metrics provide a comprehensive view of circuit quality, measuring both faithfulness to the original model and preservation of task-specific capabilities while quantifying sparsity at comparable granularities across methods.

\section{Results}
\label{sec:results}

\begin{table*}[t]
\centering
\small
\begin{tabular}{l l cccccc| ccc}
\toprule
\multirow{2}{*}{\textbf{Dataset}} & \multirow{2}{*}{\textbf{Method}} & \textbf{AB} & \textbf{AH}  & \textbf{MLPs}  & \textbf{AN}  & \textbf{M1}  & \textbf{M2} & \textbf{KL} & \textbf{LD} & \textbf{Acc.}  \\
 & & (\%) $\uparrow$  & (\%) $\uparrow$ & (\%) $\uparrow$ & (\%) $\uparrow$ & (\%) $\uparrow$& (\%) $\uparrow$ & $\downarrow$ & $\uparrow$ & (\%) $\uparrow$ \\
\midrule
\multirow{3}{*}{IOI} 
    & EAP    & 0.00 & 19.44 & 0.00 & 19.44 & 0.00 & 0.00 & 2.447 & -0.181 & 0.000 \\
    & EP     & 0.00 & 43.05  & 0.00 & 43.05 & 0.00 & 0.00 & \textbf{0.169} & 3.618 & 0.952 \\
    & DiscoGP     & 0.00 & 59.03 & 0.00  & 79.44 & 0.00 & 0.10 & 6.687 & \textbf{8.750} & 0.931 \\
    & \textbf{Ours}   & \textbf{50.00} & \textbf{76.40}  & \textbf{50.00} & \textbf{84.60} & \textbf{73.60} & \textbf{59.90} & 0.466 & 5.012 & \textbf{0.958} \\
\midrule
\multirow{3}{*}{GP}  
    & EAP    & 0.00 & 26.388 & 0.00 & 26.388 & 0.00 & 0.00 & 0.148 & 2.564 & 0.903 \\
    & EP     & 0.00 & 72.22  & 0.00 & 72.22 & 0.00 & 0.00 & \textbf{0.070} & 3.983 & \textbf{0.988} \\
    & DiscoGP     & 0.00 & 13.89  & 0.00 & 13.91 & 0.00 & 0.001 & 5.090 & \textbf{5.375} & 0.952 \\
    & \textbf{Ours}   & \textbf{16.7} & \textbf{77.8}  & \textbf{75.0} & \textbf{86.5} & \textbf{91.1} & \textbf{86.3} & 0.337 & 4.735 & 0.984 \\
\midrule
\multirow{3}{*}{GT}  
    & EAP    & 0.00 & 21.52 & 0.00 & 21.52 & 0.00 & 0.00 & 0.086 & 0.374 & 0.971 \\
    & EP     & 0.00 & 75.00  & 0.00 & 75.00 & 0.00 & 0.00 & \textbf{0.044} & \textbf{0.395} & 0.970 \\
    & DiscoGP     & 0.08 & 70.83  & 16.66 & 82.33 & 16.67 & 16.72 & 2.812 & 0.018 & 0.615 \\
    & \textbf{Ours}   & \textbf{58.30} & \textbf{93.80}  & \textbf{50.00} & \textbf{95.20} & \textbf{90.9} & \textbf{83.70} & 0.048 & 0.371 & \textbf{0.992} \\
\bottomrule
\end{tabular}
\caption{Comparison of structural sparsity and fidelity metrics on GPT-2 model for Indirect Object Identification (IOI), Gendered Pronouns (GP), and Greater Than (GT) tasks. Metrics include the pruning rates for 
Attention Blocks (AB), Attention Heads (AH), MLP blocks, and individual neurons within Attention (AN) and MLP layers (M1 and M2), as well as KL Divergence, Logit Difference, and Accuracy. }
\label{tab:structural-comparison}
\vspace{-8pt}
\end{table*}

We compare our approach against edge-pruning baselines (EAP, EP) and a recent edge and unstructured weights pruning method (DiscoGP). While \textbf{DiscoGP utilizes zero ablation rather than an adversarial stream}, limiting direct comparability, we include it for completeness. Detailed tables of discovered circuits are provided in Appendix~\ref{sec:edge_details}. Additional results on GPT-XL are provided in Appendix~\ref{sec:gpt2-xl}.

\subsection{Quantitative Performance and Sparsity}
Table~\ref{tab:structural-comparison} presents a comparative analysis of circuit sparsity and task performance. Edge sparsity and details are provided in Appendix~\ref{sec:detailed_gpt_results}. Overall, we find that our framework outperforms all of the baseline discovery methods in sparsification and compactness of the discovered circuits, while retaining high functional fidelity.


While attention-head compression is broadly comparable across methods, with our approach consistently achieving the highest sparsity, requiring as few as 9 heads (GT task), the most striking difference emerges in MLP pruning. Our framework uniquely achieves block-level MLP sparsity, a structure that competing methods fail to capture. EAP and EP retain all MLP blocks across every task, and DiscoGP achieves only limited compression on a single task. This precision extends to the neuron level, where our method prunes the vast majority of MLP neurons while all baselines, including DiscoGP's unstructured weight pruning, induce effectively zero neuron-level sparsity. The results show that our method drastically reduces the number of retained nodes and is competitive in edge compression(Appendix~\ref{sec:detailed_gpt_results}~Table~\ref{tab:structural-comparison-appendix}) with dedicated edge pruning baselines, even under a conservative fully-connected assumption for edge counting.

\subsection{Task-specific topological analysis}
Appendix~\ref{sec:edge_details} Tables~\ref{tab:pruning_summary_ioi}–\ref{tab:pruning_summary_GT} summarize node-level pruning for three GPT-2 tasks, revealing sharply different computational structures rather than uniform sparsity.
IOI is dominated by MLP-based computation. Across several layers (e.g., 1, 6, 8, 11), the model retains substantial MLPs while attention is entirely pruned, suggesting that IOI-critical transformations are implemented primarily through nonlinear residual stream updates rather than attention mechanisms.
GP exhibits a highly localized and discontinuous circuit. Layers~1 and~2 are
fully inactive, with MLP computation restricted to just three layers (0, 3,
and~7). The remaining layers retain attention heads but no MLPs, suggesting
that most of the network contributes only through sparse attention-based
processing, with nonlinear MLP computation concentrated in a few select layers.
GT displays the most extreme sparsity, with a contiguous block of inactive layers (2–5). Computation is anchored by a large MLP in Layer 0 and resumes late in Layers 8–11, consistent with a skip-like mechanism where early numerical features propagate forward with minimal intermediate processing.
Overall, node pruning exposes qualitatively different circuit topologies, distributed MLP computation (IOI), late localized processing (GP), and skip-style execution (GT).


\subsection{Results on Llama Models} Since prior approaches are computationally expensive, we include results on larger models only for our approach.  Summary of the results for Llama-3.2-1B, and Llama3.1-8B, on IOI and GP tasks are provided in Table~\ref{tab:structural-comparison-llama}. Detailed pruning results are provided in Tables~\ref{tab:pruning_summary_llama_1b_ioi}-~\ref{tab:pruning_summary_llama_8b_gp}.

\begin{table*}[t]
\centering
\small
\begin{tabular}{l l cccccc | ccc}
\toprule
\multirow{2}{*}{\textbf{Dataset}} & \multirow{2}{*}{\textbf{Model}} & \textbf{AB} & \textbf{AH}  & \textbf{MLPs}  & \textbf{AN}  & \textbf{M1}  & \textbf{M2} & \textbf{KL} & \textbf{LD} & \textbf{Acc}  \\
 & & (\%) $\uparrow$ & (\%) $\uparrow$ & (\%) $\uparrow$ & (\%) $\uparrow$ & (\%) $\uparrow$& (\%) $\uparrow$ & $\downarrow$ & $\uparrow$ & $\uparrow$ \\
\midrule
\multirow{2}{*}{IOI} 
    & 1B   & 43.8 & 88.7  & 25.0 & 92.4 & 82.3 & 76.5 & 0.38 & 5.02 & 0.96 \\
    & 8B   &56.2 & 91.0  & 28.1 & 93.9 & 79.6 & 65.1 & 0.35 & 4.66 & 0.96 \\
\midrule
\multirow{2}{*}{GP}  
    & 1B   & 62.5 & 93.8  & 37.5 & 95.7& 80.7 & 74.0 & 0.16 & 5.78 & 0.99\\
    & 8B   & 75.0 & 94.5  & 40.6 & 96.3 & 80.9 & 72.5 & 0.23 & 5.79 & 0.99 \\
\bottomrule
\end{tabular}
\caption{Comparison of structural sparsity on Llama (3.2-1B/3.1-8B) models for Indirect Object Identification (IOI), Gendered Pronouns (GP). Metrics include the pruning rates for Attention Blocks (AB), Attention Heads (AH), MLP blocks, and individual neurons within Attention (AN) and MLP layers (M1 and M2), as well as KL Divergence, Logit Difference, and Accuracy. }
\label{tab:structural-comparison-llama}
\vspace{-8pt}
\end{table*}

The results indicate that the pruned Llama circuits are highly sparse, with clear task-dependent differences: GP is consistently more compressible than IOI, especially in the larger model, suggesting that its behavior can be supported by a smaller effective subnetwork. In contrast, IOI appears less uniformly prunable, with pruning being much stronger for attention-related structures than for MLP blocks, which points to a more selective but less globally sparse circuit. Across both tasks, the larger model generally supports equally strong or stronger structural pruning while still preserving close agreement with the original model, implying that increased scale introduces redundancy rather than requiring denser task-specific circuits. 

\textbf{Qualitative Analysis.}
We analyze the discovered IOI circuit on Llama~3.2-1B in detail in Appendix~\ref{sec:Qualitative-Analysis-Llama}. Attention-score and Direct Logit Attribution analysis reveal that surviving heads in late layers implement functionally distinct roles: IO-Movers, S-Inhibitors, and S-Promoters. The findings mirror the roles identified in the manually discovered GPT-2 IOI circuit~\citep{wang2022interpretability}, while early layers contribute primarily through MLPs.

\textbf{Circuit Consistency.}
We evaluate the stability of discovered circuits across five seeds for the mask parameters on Llama~3.2-1B (Appendix~\ref{sec:seeds}). Fidelity metrics, per-layer sparsity profiles, and attention head selections remain highly consistent across seeds, indicating that our framework consistently recovers the same circuits.

\begin{table}[t]
\centering
\caption{Comparison of wall clock time and peak memory usage.}
\label{tab:efficiency}
\begin{tabular}{lcc}
\toprule
Method & Wall Clock Time (s) & Memory (MB) \\
\midrule
Ours           & 350  & 6,270  \\
EAP            & 21   & 72,794 \\
Edge-Pruning   & 2,756 & 33,354 \\
DiscoGP        & 2,926 & 17,633 \\
\bottomrule
\end{tabular}
\vspace{-8pt}
\end{table}
\subsection{Compute Requirements}
 Our node pruning method is highly computationally efficient; it requires substantially less computation than the baseline methods. It adds and trains only 55,465 additional parameters in GPT-2 small, which has 124.5M parameters. Since we have two instances of the model loaded in memory, one base model and one prunable model for KL-loss training. Our method requires the least GPU memory among all approaches.
In comparison, EAP requires over an order of magnitude more memory, because it loads the complete dataset as a single batch, and EP requires roughly 5$\times$ more, since both store all internal representations in memory. DiscoGP uses less memory than the edge-pruning methods since it does not maintain a corrupted stream, but still requires nearly 3$\times$ more than our approach.

In terms of wall clock time, our method completes training in under 6 minutes, making it substantially faster than both Edge-Pruning and DiscoGP, which each require close to 50 minutes. EAP is the fastest method in wall clock time due to its single gradient pass design, but this comes at the cost of extremely high memory usage, and fails to find small or accurate circuits compared to other baselines. Our method offers the best overall efficiency trade-off, combining low memory footprint with competitive training time.

\section{Conclusion}

We introduced a multi-granularity node-pruning framework for causal circuit discovery that simultaneously prunes transformer blocks, attention heads, MLP blocks, and individual neurons within a single optimization step. Unlike edge-pruning methods, which scale quadratically and require storing intermediate activations, our approach operates directly on nodes, achieving $3$--$11\times$ lower memory usage while discovering substantially more compact circuits.

Across three benchmarks on GPT-2, our method identifies sparser circuits at every granularity level, with particularly large gains in MLP and neuron-level compression, granularities where existing baselines achieve little to no pruning. The discovered circuits reveal task-dependent topologies: distributed MLP computation for IOI, localized late-layer processing for GP, and skiping layer execution for GT. We further demonstrate scalability to Llama~(3.2-1B/3.1-8B) on a single 40GB GPU, finding that larger models support equally
strong or stronger pruning.

An exciting future direction is combining node and edge pruning hierarchically: first selecting critical nodes via our efficient method, then recovering the interaction structure among surviving components. Such a hybrid approach may yield even smaller circuits with richer mechanistic insight.



\bibliography{colm2026_conference}
\bibliographystyle{colm2026_conference}

\include{appendix}

\end{document}

%% file: appendix.tex
\appendix

\section{Limitations}
\label{sec:limitations}

Our framework discovers node-level circuits but does not explicitly recover the interaction structure between nodes. In particular, the binary masks identify which blocks/heads/neurons are necessary, yet they do not reveal which active nodes exchange information with which others, nor the directionality or multiplicity of those interactions. Developing a hybrid node and edge procedure that infers interaction structure on top of node selections would alleviate these concerns.

\section{Potential Risks}
Targeted removal of nodes may inadvertently disable moderation, refusal, or safety-related pathways, enabling circumvention of guardrails and jailbreak-style behavior.

\section{LLM Usage}
We used an LLM-based writing assistant to improve the final draft of this manuscript. Specifically, it was used to improve clarity and polish writing. All technical content, experimental design, results, and scientific claims are the sole work of the authors.

\section{Task Details}
\label{sec:task_details}
We evaluate our method on three circuit-discovery tasks that test different computational capabilities of language models. 

\textbf{Greater-Than (GT).} This task probes the numerical reasoning capabilities of language models, specifically their ability to perform temporal comparisons between years. For example, given a prompt \textit{"The war lasted from the year 1743 to the year 17\_\_"}, the model should assign higher probability to completions to numerics 44-99 than to 00-42. 200 examples were used for training, 200 for validation, and 1000 for testing.

We utilize the dataset from \cite{hanna2023does}, which consists of 5 templates, 120 noun choices, and years spanning 1100-2199. The dataset consists of 12,540 samples that are divided into train, validation and test splits. This task requires models to extract the starting year, understand that ending years must be chronologically greater than starting years, and map this constraint to select valid two-digit completions.

\textbf{Indirect Object Identification (IOI).} The IOI task tests syntactic tracking and entity binding. For example, given a sentence \textit{"Friends Juana and Kristi found a mango at the bar. Kristi gave it to"}, the model is expected to predict “Juana” as the recipient. 

We evaluate IOI dataset, which incorporates 30 diverse syntactic templates comprising 200 examples each for training and validation, and  1000 test instances. Example templates include constructions such as “Then, B and A had a long argument. Afterwards, B said to → A". The task requires the model to track entity positions, suppress repeated names, and correctly resolve the target entity for output. Prior work \cite{conmy2023towards} has shown that the IOI circuit spans multiple layers, involving specialized components responsible for name identification, positional encoding, and duplicate name inhibition.

\textbf{Gendered Pronouns (GP).} This task probes learned associations between names and pronouns. For example, given the prompt \textit{"So Evan is a really great friend, isn't "}, the model should assign a higher probability to the gender-consistent pronoun “he” rather than the incorrect alternative “she.”. 

The dataset used is constructed by incorporating the top 1,000 most popular male and female baby names from 2000, creating 150 train/validation examples each and 378 test examples. Unlike IOI, which evaluates syntactic reasoning, this task probes whether language models encode societal gender associations linked to names. Circuit analysis \cite{conmy2023towards} indicates that early layers are responsible for encoding gender-related features, while middle layers bind these features to the corresponding gendered pronouns.

\section{Circuit Discovery Algorithm:}
Algorithm~\ref{alg:node_pruning} provides an overview of the circuit discovery algorithm.

\begin{algorithm*}[t]
\footnotesize
\caption{Multi-Granularity Circuit Discovery}
\begin{algorithmic}[1]

\Statex \textbf{Training Phase}
\Require Model $f_\theta$ with layers $l \in L$, task dataset $\mathcal{D}$, corruption function $C$
\State Initialize Hard-Concrete mask parameters $ \alpha_{i}$ for all granularities $\mathcal{G}$ across the Model

\For{$\text{epoch} \gets 1$ \textbf{to} $N_{\text{epochs}}$}
  \For{batch $(x, y)$ \textbf{in} $\mathcal{D}$} 
    \State $x_{\text{corrupt}} \gets C(x)$
    
    \State  $M \sim \mathrm{H}(\log \alpha_{i}, \beta)$ \Comment{Sample masks (Equation~\ref{eq:hard_concrete_mask})}
    
    \State $h_L^{\text{base}} \gets f_{\theta}(x)$

    \For{$l \gets 1$ \textbf{ to } $L$} \Comment{For each layer in $f_{\theta}$ and across all nodes using $m_l = \{m_{l,i}\}$}
      \State $h_{l}^{\text{clean}} \gets f_l(\mathrm{I}\!\left(h_{l-1}^{\text{clean}}, h_{l-1}^{\text{corrupt}}, m_{l}\right))$ \Comment{Interpolation function I (Equation~\ref{eq:forward_pass})}
      \State $h_{l}^{\text{corrupt}} \gets  f_l(h_{l-1}^{\text{corrupt}}) $
    \EndFor

    \State $\mathcal{L} = (1-\lambda)(\mathcal{L}_{\text{faith}}(h^{\text{circuit}},h^{\text{base}})
+ \mathcal{L}_{\text{task}}(h_{gt}^{\text{circuit}},h_{distr}^{\text{circuit}})) + \lambda \sum_{i=1}^{N} \mathcal{L}_0(m_i)$

  \EndFor
\EndFor
\end{algorithmic}

\label{alg:node_pruning}
\end{algorithm*}
\section{Training Details:}
The number of training samples utilized for the Gender Pronouns and Greater than tasks is 150. For the Indirect Object Identification task, 200 training samples are used. For circuit extraction, a batch size of 32 with 500 epochs is used. For all tasks, we fix the max sequence length to 64. Adam optimizer~\citep{kingma2015adam} is used with a learning rate $3e-2$.

\section{Compute Resources}
Experiments are conducted using an Nvidia A100-40GB GPU. Some baselines require larger V-RAM, for which we use Nvidia H100-96 GB.

\section{Hyper Parameters Details:}

Our approach utilizes a single hyperparameter: the sparsity coefficient $\lambda$. To ensure consistent sparsity pressure across granularities, we normalize the sparsity loss within each granularity to $[0, 1]$ and then renormalize across all granularities, so that $\lambda$ controls overall sparsity uniformly. We select the sparsity scale $\lambda$ per task and model via a small number of preliminary runs. $\lambda$ values used are [0.7-0.95], depending upon task and model.

\section{Consistency of the Discovered circuits}
\label{sec:seeds}

\begin{figure}[h]
  \centering
  \includegraphics[width=1.0\textwidth]{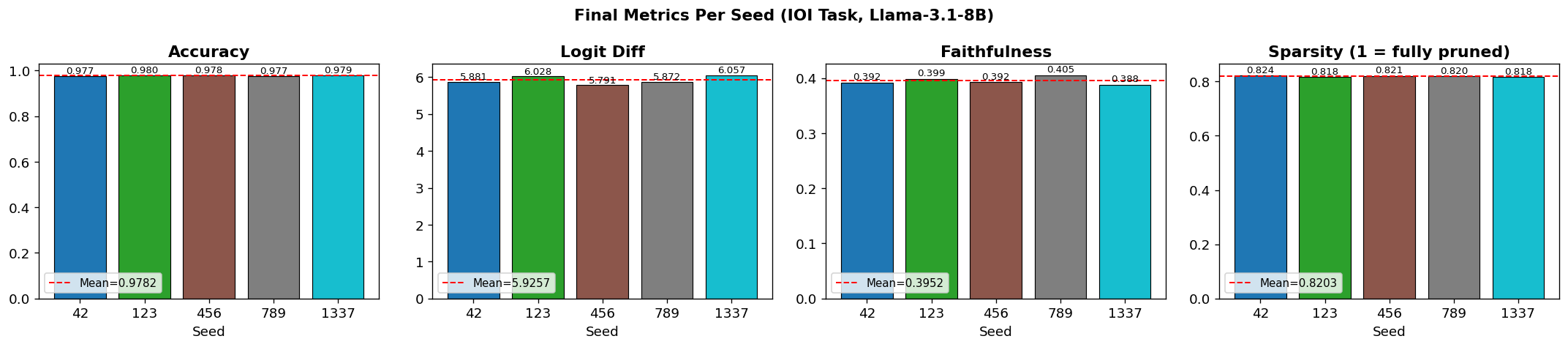}
  \caption{Metrics across training seeds for circuit extraction.}
  \label{fig:seed_metrics}
\end{figure}

\begin{figure}[h]
  \centering
  \includegraphics[width=1.0\textwidth]{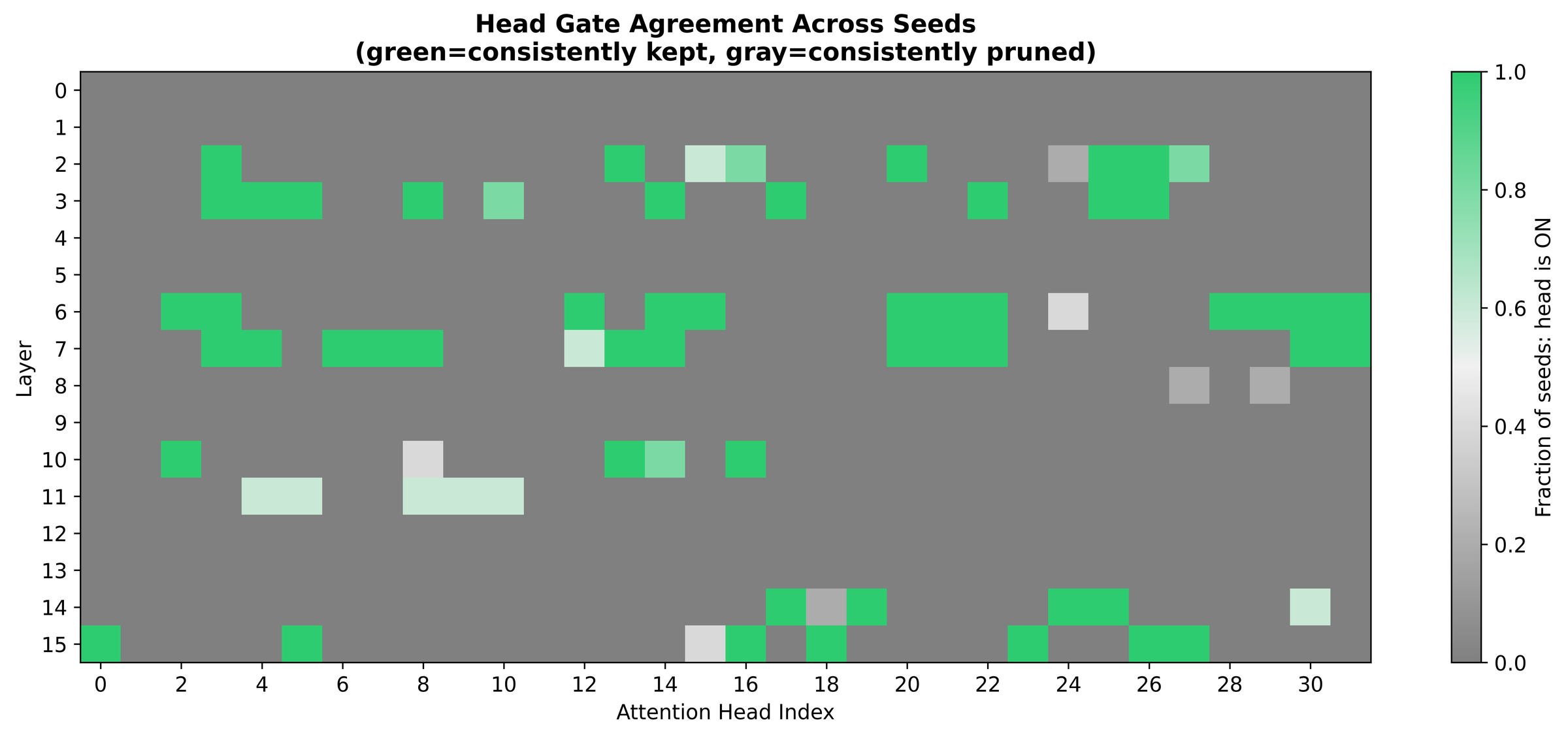}
  \caption{Surviving attention heads in the pruned Llama~3.2-1B circuit for the IOI task. Green cells indicate heads retained across all five random seeds; Gray cells indicate consistently pruned heads.}
  \label{fig:seed_heads}
\end{figure}

\begin{figure}[h]
  \centering
  \includegraphics[width=1.0\textwidth]{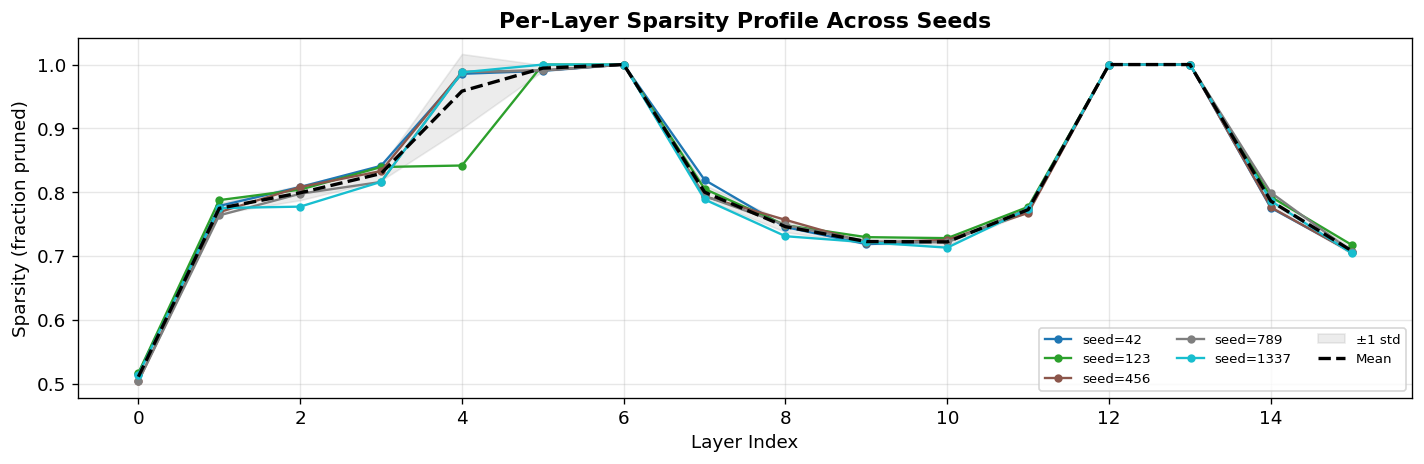}
  \caption{Compressed circuit structure for Llama~3.2-1B on the indirect object identification task. Active attention blocks (blue), MLP blocks, and their retained neuron counts are shown per layer. Layers~5-6 and~12-13 are fully pruned.}
  \label{fig:seed_layers}
\end{figure}

To test how consistent and stable discovered circuits are, we analyse the discovered circuits and fidelity metrics across 5 seed initializations for the mask parameters for the IOI task on Llama-3.2-1B model. Figure~\ref{fig:seed_metrics} provides fidelity and sparsity metrics across different seeds. Figure~\ref{fig:seed_heads} shows attention head heat maps across seeds. Figure~\ref{fig:seed_layers} provides sparcities across layers across different seeds. Across different seeds, we find highly consistent behaviour of the pruning profiles.

\section{Compression Results GPT-2}
\label{sec:detailed_gpt_results}
Table~\ref{tab:structural-comparison-appendix} presents structural details of compression metrics, along with edge compression ratios on GPT-2 small model across all three tasks.

\begin{table*}[t]
\centering
\small
\begin{tabular}{l l cccccc c}
\toprule
\multirow{2}{*}{\textbf{Dataset}} & \multirow{2}{*}{\textbf{Method}} & \textbf{AB} & \textbf{AH}  & \textbf{MLPs}  & \textbf{AN}  & \textbf{M1 N}  & \textbf{M2 N} & \textbf{Edge Comp.}  \\
 & & (\%) $\uparrow$  & (\%) $\uparrow$ & (\%) $\uparrow$ & (\%) $\uparrow$ & (\%) $\uparrow$& (\%) $\uparrow$ & (\%) $\uparrow$ \\
\midrule
\multirow{3}{*}{IOI} 
    & EAP    & 0.00 & 19.44 & 0.00 & 19.44 & 0.00 & 0.00 & 96.74 \\
    & EP     & 0.00 & 43.05  & 0.00 & 43.05 & 0.00 & 0.00 & 94.77 \\
    & DiscoGP     & 0.00 & 59.03 & 0.00  & 79.44 & 0.00 & 0.10 & \textbf{98.49} \\
    & \textbf{Ours}   & \textbf{50.0} & \textbf{76.4}  & \textbf{50.0} & \textbf{84.6} & \textbf{73.6} & \textbf{59.9} & 93.73 \\
\midrule
\multirow{3}{*}{GP}  
    & EAP    & 0.00 & 26.388 & 0.00 & 26.388 & 0.00 & 0.00 & 93.74 \\
    & EP     & 0.00 & 72.22  & 0.00 & 72.22 & 0.00 & 0.00 & \textbf{98.88} \\
    & DiscoGP     & 0.00 & 13.89  & 0.00 & 13.91 & 0.00 & 0.001 & 95.76 \\
    & \textbf{Ours}   & \textbf{16.7} & \textbf{77.8}  & \textbf{75.0} & \textbf{86.5} & \textbf{91.1} & \textbf{86.3} & 94.69 \\
\midrule
\multirow{3}{*}{GT}  
    & EAP    & 0.00 & 21.52 & 0.00 & 21.52 & 0.00 & 0.00 & 95.95 \\
    & EP     & 0.00 & 75.00  & 0.00 & 75.00 & 0.00 & 0.00 & 98.75 \\
    & DiscoGP     & 0.08 & 70.83  & 16.66 & 82.33 & 16.67 & 16.72 & 98.60 \\
    & \textbf{Ours}   & \textbf{58.3} & \textbf{93.8}  & \textbf{50.0} & \textbf{95.2} & \textbf{90.9} & \textbf{83.7} & \textbf{99.21} \\
\bottomrule
\end{tabular}
\caption{Quantitative comparison of structural sparsity for Indirect Object Identification (IOI), Gendered Pronouns (GP), and Greater Than (GT) tasks. Metrics include percentage of edges compressed (Edge Comp.) and the pruning rates for Attention Blocks (AB), Attention Heads (AH), MLP blocks, and individual neurons within Attention (AN) and MLP layers (M1 N and M2 N).}
\label{tab:structural-comparison-appendix}
\vspace{-8pt}
\end{table*}

\section{Circuit detailed Summaries}
\label{sec:edge_details}
Across tasks, the node pruning summaries (Tables~\ref{tab:pruning_summary_ioi}–\ref{tab:pruning_summary_GT}) reveal highly structured sparsity: large contiguous bands of fully pruned layers or blocks, with activity concentrating in a few late layers. For \textit{Greater Than}, many early–mid layers disable entirely, while attention and MLP usage peaks in layers 9–10. For \textit{Gender Pronouns}, the mid-stack is largely inactive, punctuated by bursts of attention activity (e.g., layers 5, 10–12) and sparse MLP use. For \textit{IOI}, attention is mostly pruned except in the final layers, whereas MLP blocks remain active across nearly the whole depth with substantial hidden/output retention, indicating MLP-dominant computation. In contrast, the edge-based summaries (EP and EAP; Tables~\ref{tab:ioi_ep_ea} – \ref{tab:gt_ep_ea}) keep all layers active and vary primarily in the number of retained heads, with EAP consistently preserving more heads than EP. Overall, node-level results exhibit modular, layer-selective circuits, while edge-based results favor distributed connectivity with minimal layer-level shutoff. DiscoGP have similar results in Tables~\ref{tab:ioi_disco}-\ref{tab:gt_disco}.



\begin{table*}[ht]
\centering
\small

\renewcommand{\arraystretch}{1.2}
\begin{tabular}{c|c|c|c|c|c|c}
\toprule
\textbf{Layer} & \textbf{Attn Block} & \textbf{MLP Block} & \textbf{Attn Heads} & \textbf{Attn Neurons} & \textbf{MLP Hidden} & \textbf{MLP Output} \\
\midrule
0  & Active & Active & 9/12  & 299/768 & 2120/3072 & 732/768 \\
1  & Pruned & Active & 0/12  & 0/768   & 1588/3072 & 635/768 \\
2  & Pruned & Pruned & 0/12  & 0/768   & 0/3072    & 0/768   \\
3  & Active & Pruned & 6/12  & 272/768 & 0/3072    & 0/768   \\
4  & Pruned & Pruned & 0/12  & 0/768   & 0/3072    & 0/768   \\
5  & Active & Active & 3/12  & 138/768 & 1513/3072 & 604/768 \\
6  & Pruned & Active & 0/12  & 0/768   & 1495/3072 & 595/768 \\
7  & Active & Pruned & 5/12  & 247/768 & 0/3072    & 0/768   \\
8  & Pruned & Active & 0/12  & 0/768   & 1454/3072 & 562/768 \\
9  & Active & Pruned & 4/12  & 189/768 & 0/3072    & 0/768   \\
10 & Active & Pruned & 7/12  & 276/768 & 0/3072    & 0/768   \\
11 & Pruned & Active & 0/12  & 0/768   & 1565/3072 & 571/768 \\
\bottomrule

\end{tabular}
\caption{Node Pruning(Ours) summary for Indirect Object Identification GPT-2 Model: attention blocks, MLP blocks, and retained components.}
\label{tab:pruning_summary_ioi}
\end{table*}

\begin{table*}[ht]
\centering
\small

\renewcommand{\arraystretch}{1.2}
\begin{tabular}{c|c|c|c|c|c|c}
\toprule
\textbf{Layer} & \textbf{Attn Block} & \textbf{MLP Block} & \textbf{Attn Heads} & \textbf{Attn Neurons} & \textbf{MLP Hidden} & \textbf{MLP Output} \\
\midrule
0 & Active & Active & 9/12 & 311/768 & 1682/3072 & 579/768 \\
1 & Pruned & Pruned & 0/12 & 0/768 & 0/3072 & 0/768 \\
2 & Pruned & Pruned & 0/12 & 0/768 & 0/3072 & 0/768 \\
3 & Active & Active & 3/12 & 126/768 & 1054/3072 & 319/768 \\
4 & Active & Pruned & 2/12 & 88/768 & 0/3072 & 0/768 \\
5 & Active & Pruned & 3/12 & 120/768 & 0/3072 & 0/768 \\
6 & Active & Pruned & 3/12 & 119/768 & 0/3072 & 0/768 \\
7 & Active & Active & 3/12 & 118/768 & 555/3072 & 363/768 \\
8 & Active & Pruned & 3/12 & 106/768 & 0/3072 & 0/768 \\
9 & Active & Pruned & 2/12 & 78/768 & 0/3072 & 0/768 \\
10 & Active & Pruned & 1/12 & 59/768 & 0/3072 & 0/768 \\
11 & Active & Pruned & 3/12 & 121/768 & 0/3072 & 0/768 \\
\bottomrule
\end{tabular}

\caption{Node Pruning(Ours) summary for Gender Pronouns Task GPT-2 Model: attention blocks, MLP blocks, and retained components.}

\label{tab:pruning_summary_GP}
\end{table*}

\begin{table*}[ht]
\centering
\small

\renewcommand{\arraystretch}{1.2}
\begin{tabular}{c|c|c|c|c|c|c}
\toprule
\textbf{Layer} & \textbf{Attn Block} & \textbf{MLP Block} & \textbf{Attn Heads} & \textbf{Attn Neurons} & \textbf{MLP Hidden} & \textbf{MLP Output} \\
\midrule
0 & Active & Active & 3/12 & 144/768 & 895/3072 & 440/768 \\
1 & Pruned & Active & 0/12 & 0/768 & 569/3072 & 155/768 \\
2 & Pruned & Pruned & 0/12 & 0/768 & 0/3072 & 0/768 \\
3 & Pruned & Pruned & 0/12 & 0/768 & 0/3072 & 0/768 \\
4 & Pruned & Pruned & 0/12 & 0/768 & 0/3072 & 0/768 \\
5 & Pruned & Pruned & 0/12 & 0/768 & 0/3072 & 0/768 \\
6 & Active & Pruned & 2/12 & 92/768 & 0/3072 & 0/768 \\
7 & Active & Pruned & 1/12 & 53/768 & 0/3072 & 0/768 \\
8 & Active & Active & 2/12 & 90/768 & 382/3072 & 130/768 \\
9 & Active & Active & 1/12 & 60/768 & 481/3072 & 335/768 \\
10 & Pruned & Active & 0/12 & 0/768 & 619/3072 & 338/768 \\
11 & Pruned & Active & 0/12 & 0/768 & 420/3072 & 108/768 \\
\bottomrule
\end{tabular}
\caption{Node Pruning(Ours) summary for Greater Than Task GPT-2 Model: attention blocks, MLP blocks, and retained components.}
\label{tab:pruning_summary_GT}
\end{table*}

\begin{table*}[ht]
\small
\centering
\begin{tabular}{c|c|c||c|c|c}
\toprule
\multicolumn{3}{c||}{\textbf{EP}} & \multicolumn{3}{c}{\textbf{EAP}} \\
\midrule
\textbf{Layer} & \textbf{Attn Heads Active} & \textbf{MLP Block} 
& \textbf{Layer} & \textbf{Attn Heads Active} & \textbf{MLP Block} \\
\midrule
0  &  12 / 12 & Active & 0  & 12 / 12 & Active \\
1  &  5 / 12 & Active & 1  & 11 / 12 & Active \\
2  &  8 / 12 & Active & 2  & 10 / 12 & Active \\
3  &  6 / 12 & Active & 3  & 10 / 12 & Active \\
4  &  4 / 12 & Active & 4  &  8 / 12 & Active \\
5  &  5 / 12 & Active & 5  & 11 / 12 & Active \\
6  &  6 / 12 & Active & 6  & 12 / 12 & Active \\
7  &  5 / 12 & Active & 7  &  9 / 12 & Active \\
8  &  5 / 12 & Active & 8  &  6 / 12 & Active \\
9  &  6 / 12 & Active & 9  & 10 / 12 & Active \\
10 &  9 / 12 & Active & 10 &  9 / 12 & Active \\
11 &  11 / 12 & Active & 11 &  8 / 12 & Active \\
\bottomrule
\end{tabular}
\caption{Indirect Object Identification Task: Pruning Summary for Edge Pruning (EP) vs. Edge Attribution Patching (EAP)}
\label{tab:ioi_ep_ea}
\end{table*}

\begin{table*}[ht]
\small
\centering
\begin{tabular}{c|c|c||c|c|c}
\toprule
\multicolumn{3}{c||}{\textbf{EP}} & \multicolumn{3}{c}{\textbf{EAP}} \\
\midrule
\textbf{Layer} & \textbf{Attn Heads Active} & \textbf{MLP Block} 
& \textbf{Layer} & \textbf{Attn Heads Active} & \textbf{MLP Block} \\
\midrule
0  &  8 / 12 & Active & 0  & 12 / 12 & Active \\
1  &  2 / 12 & Active & 1  & 10 / 12 & Active \\
2  &  4 / 12 & Active & 2  & 10 / 12 & Active \\
3  &  3 / 12 & Active & 3  &  9 / 12 & Active \\
4  &  1 / 12 & Active & 4  &  9 / 12 & Active \\
5  &  1 / 12 & Active & 5  &  9 / 12 & Active \\
6  &  3 / 12 & Active & 6  &  9 / 12 & Active \\
7  &  2 / 12 & Active & 7  &  8 / 12 & Active \\
8  &  5 / 12 & Active & 8  & 10 / 12 & Active \\
9  &  3 / 12 & Active & 9  &  6 / 12 & Active \\
10 &  3 / 12 & Active & 10 &  8 / 12 & Active \\
11 &  1 / 12 & Active & 11 &  6 / 12 & Active \\
\bottomrule
\end{tabular}
\caption{Gender Pronouns Task: Pruning Summary for Edge Pruning (EP) vs. Edge Attribution Patching (EAP)}
\label{tab:gp_ep_ea}
\end{table*}

\begin{table*}[ht]
\small
\centering
\begin{tabular}{c|c|c||c|c|c}
\toprule
\multicolumn{3}{c||}{\textbf{EP}} & \multicolumn{3}{c}{\textbf{EAP}} \\
\midrule
\textbf{Layer} & \textbf{Attn Heads Active} & \textbf{MLP Block} 
& \textbf{Layer} & \textbf{Attn Heads Active} & \textbf{MLP Block} \\
\midrule
0  &  9 / 12 & Active & 0  & 12 / 12 & Active \\
1  &  4 / 12 & Active & 1  & 11 / 12 & Active \\
2  &  5 / 12 & Active & 2  & 12 / 12 & Active \\
3  &  8 / 12 & Active & 3  & 12 / 12 & Active \\
4  &  6 / 12 & Active & 4  & 11 / 12 & Active \\
5  &  5 / 12 & Active & 5  & 12 / 12 & Active \\
6  &  5 / 12 & Active & 6  & 10 / 12 & Active \\
7  &  6 / 12 & Active & 7  & 11 / 12 & Active \\
8  &  8 / 12 & Active & 8  & 10 / 12 & Active \\
9  &  5 / 12 & Active & 9  &  6 / 12 & Active \\
10 &  7 / 12 & Active & 10 &  4 / 12 & Active \\
11 &  6 / 12 & Active & 11 &  2 / 12 & Active \\
\bottomrule
\end{tabular}
\caption{Greater Than Task: Pruning Summary for Edge Pruning (EP) vs. Edge Attribution Patching (EAP)}
\label{tab:gt_ep_ea}
\end{table*}

\begin{table*}[ht]
\centering
\small
\renewcommand{\arraystretch}{1.2}
\begin{tabular}{c|c|c|c|c|c|c}
\toprule
\textbf{Layer} & \textbf{Attn Block} & \textbf{MLP Block} & \textbf{Attn Heads} & \textbf{Attn Neurons} & \textbf{MLP Hidden} & \textbf{MLP Output} \\
\midrule
0 & Active & Active & 7/12 & 255/768 & 3072/3072 & 768/768 \\
1 & Active & Active & 9/12 & 276/768 & 3072/3072 & 768/768 \\
2 & Active & Active & 7/12 & 56/768 & 3072/3072 & 768/768 \\
3 & Active & Active & 1/12 & 1/768 & 3072/3072 & 768/768 \\
4 & Active & Active & 3/12 & 23/768 & 3072/3072 & 768/768 \\
5 & Active & Active & 4/12 & 18/768 & 3072/3072 & 768/768 \\
6 & Active & Active & 4/12 & 96/768 & 3072/3072 & 766/768 \\
7 & Active & Active & 4/12 & 66/768 & 3072/3072 & 765/768 \\
8 & Active & Active & 2/12 & 37/768 & 3072/3072 & 766/768 \\
9 & Active & Active & 4/12 & 198/768 & 3072/3072 & 766/768 \\
10 & Active & Active & 5/12 & 297/768 & 3072/3072 & 768/768 \\
11 & Active & Active & 9/12 & 572/768 & 3072/3072 & 768/768 \\
\bottomrule
\end{tabular}
\caption{DiscoGP summary for Indirect Object Identification GPT-2 Model: attention blocks, MLP blocks, and retained components.}
\label{tab:ioi_disco}
\end{table*}

\begin{table*}[ht]
\centering
\small
\renewcommand{\arraystretch}{1.2}
\begin{tabular}{c|c|c|c|c|c|c}
\toprule
\textbf{Layer} & \textbf{Attn Block} & \textbf{MLP Block} & \textbf{Attn Heads} & \textbf{Attn Neurons} & \textbf{MLP Hidden} & \textbf{MLP Output} \\
\midrule
0 & Active & Active & 11/12 & 704/768 & 3072/3072 & 768/768 \\
1 & Active & Active & 12/12 & 768/768 & 3072/3072 & 768/768 \\
2 & Active & Active & 11/12 & 704/768 & 3072/3072 & 768/768 \\
3 & Active & Active & 10/12 & 640/768 & 3072/3072 & 768/768 \\
4 & Active & Active & 11/12 & 702/768 & 3072/3072 & 768/768 \\
5 & Active & Active & 11/12 & 704/768 & 3072/3072 & 768/768 \\
6 & Active & Active & 12/12 & 768/768 & 3072/3072 & 768/768 \\
7 & Active & Active & 8/12 & 512/768 & 3072/3072 & 767/768 \\
8 & Active & Active & 10/12 & 640/768 & 3072/3072 & 768/768 \\
9 & Active & Active & 8/12 & 512/768 & 3072/3072 & 768/768 \\
10 & Active & Active & 8/12 & 512/768 & 3072/3072 & 768/768 \\
11 & Active & Active & 12/12 & 768/768 & 3072/3072 & 768/768 \\
\bottomrule
\end{tabular}
\caption{DiscoGP summary for Gender Pronouns GPT-2 Model: attention blocks, MLP blocks, and retained components.}
\label{tab:gp_disco}
\end{table*}

\begin{table*}[ht]
\centering
\small

\renewcommand{\arraystretch}{1.2}
\begin{tabular}{c|c|c|c|c|c|c}
\toprule
\textbf{Layer} & \textbf{Attn Block} & \textbf{MLP Block} & \textbf{Attn Heads} & \textbf{Attn Neurons} & \textbf{MLP Hidden} & \textbf{MLP Output} \\
\midrule
0 & Active & Active & 5/12 & 320/768 & 3072/3072 & 768/768 \\
1 & Active & Active & 6/12 & 357/768 & 3072/3072 & 768/768 \\
2 & Active & Active & 6/12 & 237/768 & 3072/3072 & 768/768 \\
3 & Active & Active & 6/12 & 21/768 & 3072/3072 & 768/768 \\
4 & Active & Active & 3/12 & 24/768 & 3072/3072 & 768/768 \\
5 & Active & Active & 4/12 & 61/768 & 3072/3072 & 768/768 \\
6 & Active & Active & 1/12 & 49/768 & 3072/3072 & 768/768 \\
7 & Active & Active & 1/12 & 54/768 & 3072/3072 & 766/768 \\
8 & Active & Active & 1/12 & 16/768 & 3072/3072 & 766/768 \\
9 & Pruned & Active & 0/12 & 0/768 & 3072/3072 & 767/768 \\
10 & Active & Pruned & 4/12 & 193/768 & 0/3072 & 0/768 \\
11 & Active & Pruned & 5/12 & 295/768 & 0/3072 & 0/768 \\
\bottomrule
\end{tabular}
\caption{DiscoGP summary for Greater Than GPT-2 Model: attention blocks, MLP blocks, and retained components.}
\label{tab:gt_disco}
\end{table*}

\section{Qualitative analysis Llama-3.2-1B}
\label{sec:Qualitative-Analysis-Llama}

\subsection{Surviving Attentions}
\begin{figure}[h]
    \centering
        \includegraphics[width=\textwidth]{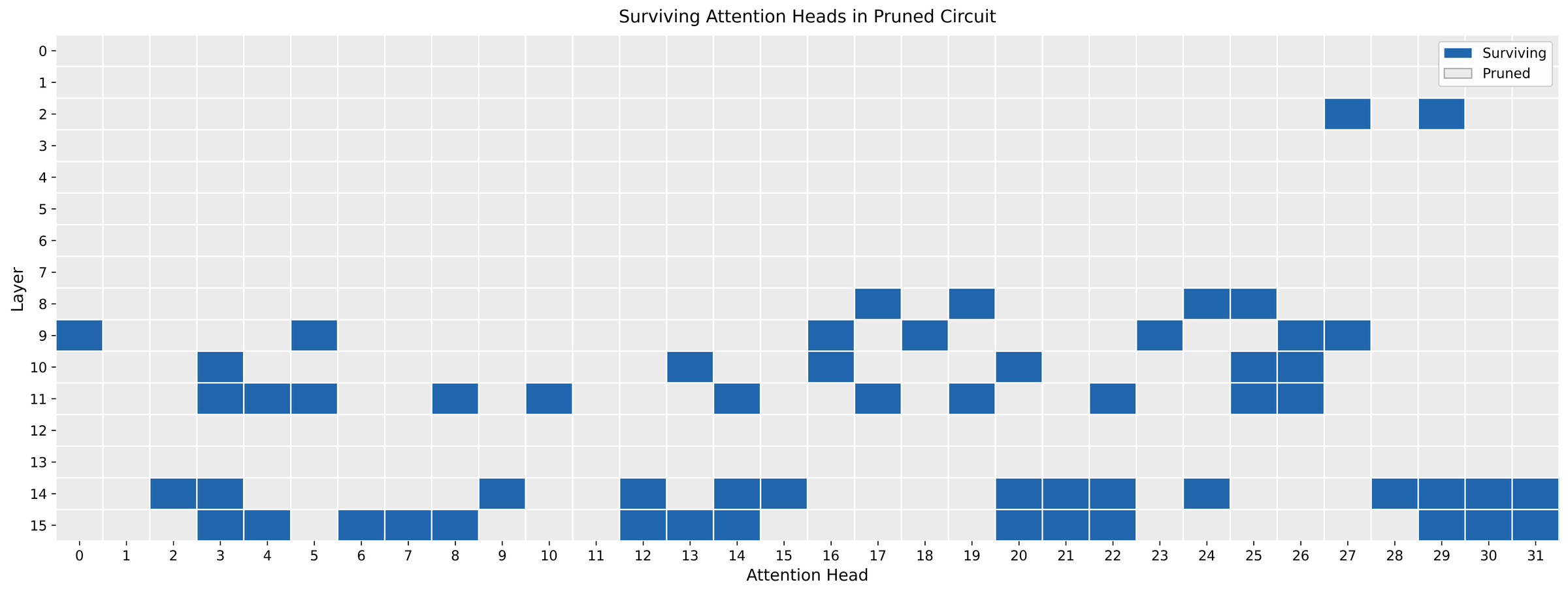}
        \label{fig:sub1}
    \caption{Llama 3.2-1B compressed circuit on indirect object identification task.}
    \label{fig:surviving_heads_llama}
\end{figure}

Figure~\ref{fig:surviving_heads_llama} visualizes the surviving attention heads in the pruned Llama~3.2-1B circuit for the IOI task. The circuit retains 58 of 512 total attention heads (11.3\%), concentrated in layers 8-11 and 14-15. Early layers (0-7) are almost entirely pruned of attention, with only layer~2 retaining 2 heads. This pattern indicates that attention-mediated computation for IOI in Llama~3.2-1B is predominantly a late-layer phenomenon. Early layers
contribute primarily through their MLP pathways, while the final layers perform the entity-tracking and name-mover operations through attention.
This is consistent with prior findings that name-mover and backup name-mover heads concentrate in the final layers of the model
\citep{wang2022interpretability}.

\subsection{Attention analysis}
\begin{figure}[h]
  \centering
  \includegraphics[width=1.0\textwidth]{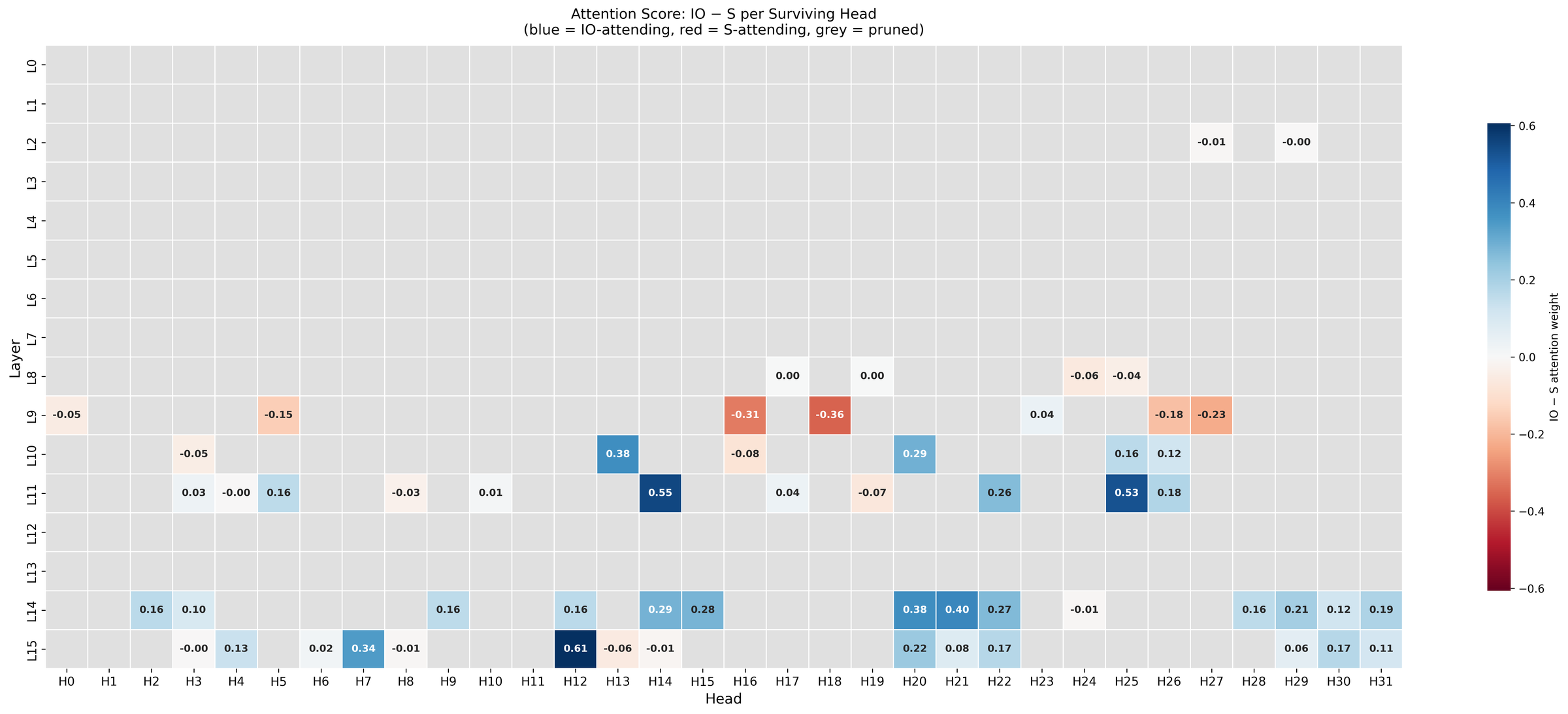}
  \caption{Attention Score analysis heat map of IO-S tokens.}
  \label{fig:attn_heatmap}
\end{figure}

To understand "what" the surviving attention heads compute, we analyze their attention patterns on IOI prompts. Figure~\ref{fig:attn_heatmap} presents the attention scores at the prediction position for the indirect object (IO) and repeated subject (S) tokens, across all surviving heads.

The analysis reveals a clear pattern where late layer heads in layers~14-15 exhibit two distinct behaviors: (1) Majority heads are~\textbf{IO-attending heads} that place high attention weight on the indirect object token at the prediction position (e.g., (layer 14, head 15), (layer 15, head 20), (layer 15, head 12)), and (2)~\textbf{S-attending heads}
that attend to the repeated subject (e.g., (layer 15, head 6), (layer 15, head 13)). Heads that attend strongly to the IO token while also exhibiting positive Direct Logit Attribution (DLA) correspond to \textbf{IO-Mover} heads, which directly promote the correct answer. Conversely, heads that attend to S but show negative DLA function as \textbf{S-Inhibitor} heads, suppressing the incorrect repeated name.

\subsection{DLA Analysis:}
We use Direct Logit Attribution (DLA) to decompose the final logit difference (IO score minus S score) into per-component contributions. DLA approximates each component's effect on the output by projecting its contribution through the unembedding matrix with frozen layer-norm statistics~\citep{elhage2021mathematical, wang2022interpretability}:
\begin{equation}
  \text{DLA}(c) = W_U \cdot \widehat{\text{LN}}(v_c),
\end{equation}
where $v_c$ is the residual-stream contribution of component $c$ and
$\widehat{\text{LN}}$ applies layer normalization with scale factors computed from the full residual stream.

\begin{figure}[h]
  \centering
  \includegraphics[width=1.0\textwidth]{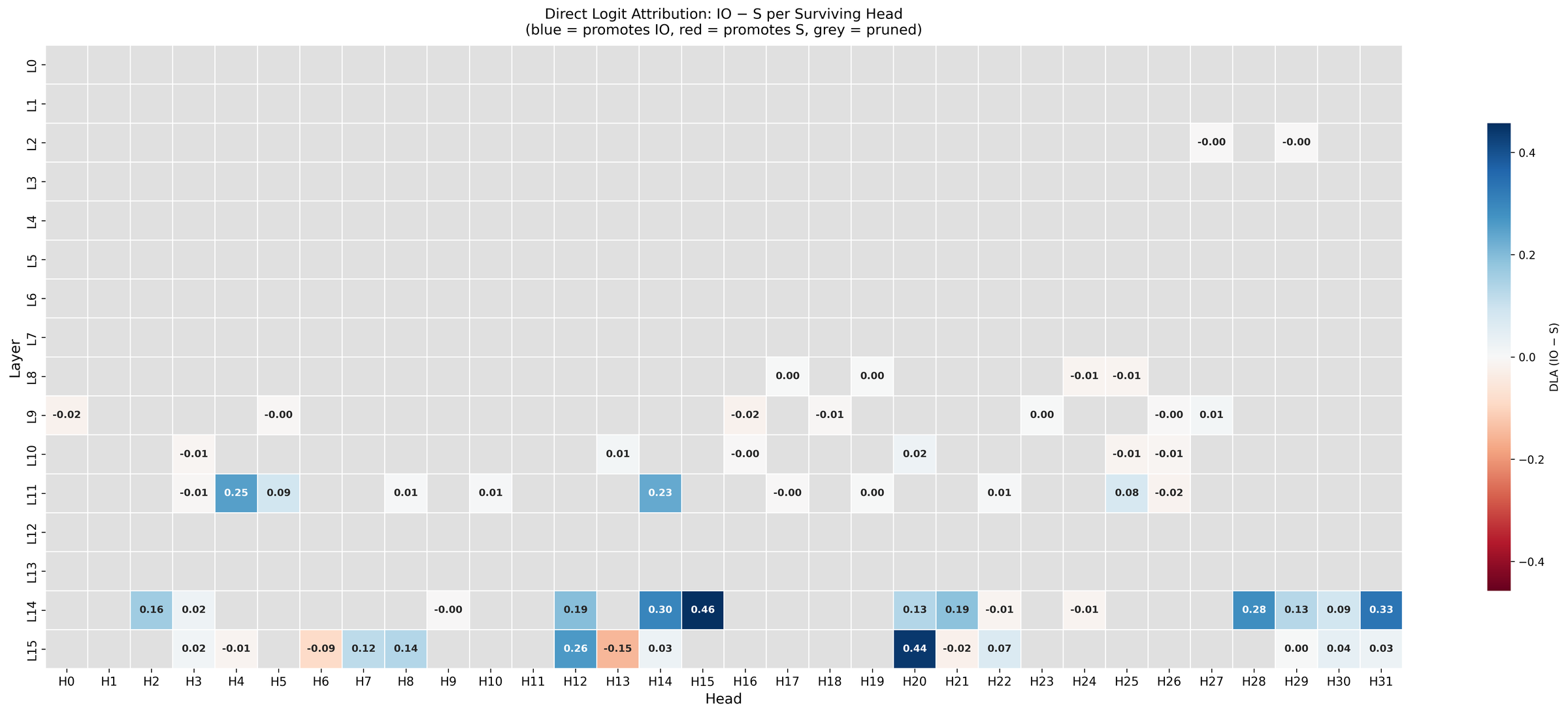}
  \caption{DLA Score Heat Map IO-S.}
  \label{fig:dla_heatmap}
\end{figure}

Figure~\ref{fig:dla_heatmap} presents the DLA scores (IO $-$ S) for all
surviving attention heads. The results confirm a clear functional pattern:

\begin{itemize}
  \item \textbf{IO-Movers} (positive DLA, high IO attention): Heads such as
    (layer 14, head 15), (layer 14, head 31), (layer 15, head 20), and (layer 15, head 12) contribute strongly to the correct
    prediction. These heads attend to the IO token and project its identity into the output logits.
  \item \textbf{Negative IO-Movers} (negative DLA, high IO attention): A small number of heads (e.g., layer 15, head 6), (layer 15, head 13)) attend to the IO token, but suppress it. These may serve as mechanisms that prevent over-confident predictions.
  \item \textbf{S-Inhibitors} (negative DLA, moderate S attention): Heads such as (layer 15, head 8) attend to the repeated subject and actively suppress its logit, complementing the IO-Movers.
  \item \textbf{S-Promoters} (positive DLA, high S attention): A few heads weakly promote the subject token, potentially reflecting residual competition or serving as backup pathways.
\end{itemize}

The DLA analysis validates that the pruned circuit retains a functionally complete set of heads: IO-Movers to promote the correct answer, S-Inhibitors to suppress the distractor, and a small set of modulatory heads. The overall pattern closely mirrors the functional roles identified in the manually discovered GPT-2 IOI circuit \citep{wang2022interpretability}, providing evidence that analogous computational structures exist in larger models.

\begin{figure}[h]
  \centering
  \includegraphics[width=1.0\textwidth]{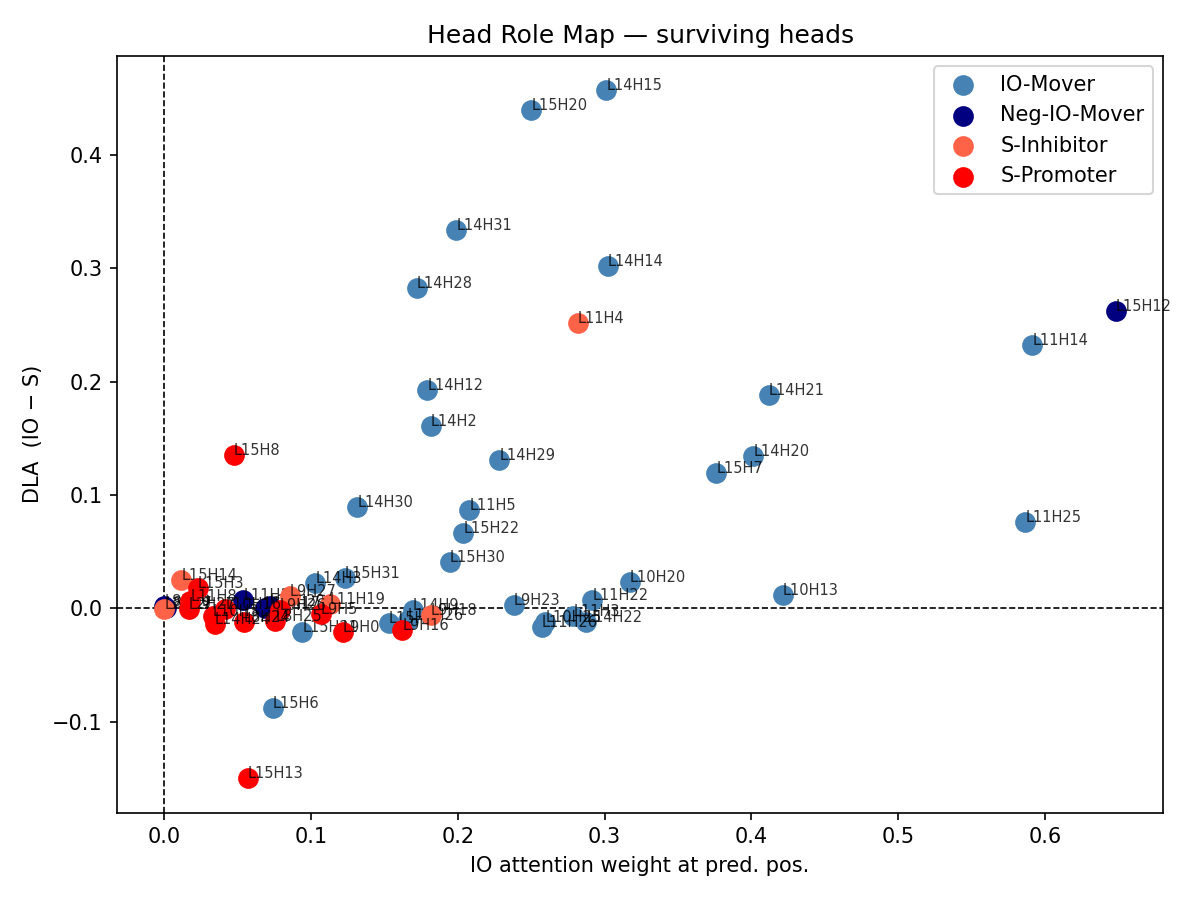}
  \caption{Head Role Map for surviving attention heads in the Llama~3.21B IOI circuit. Each head is plotted by its IO attention eight at the prediction position ($x$-axis) and its Direct Logit attribution score for IO $-$ S ($y$-axis). Colors indicate functional roles: IO-Movers (light blue), Negative IO-Movers (dark blue), S-Inhibitors (orange), and S-Promoters (red).}
  \label{fig:yourlabel}
\end{figure}


\section{Llama Detailed Results}
\label{sec:llama-appendix}

\subsection{Llama 3.2-1B}

Tables~\ref{tab:pruning_summary_llama_1b_ioi} and~\ref{tab:pruning_summary_llama_1b_gp} present the per-layer
node pruning summaries for Llama~3.2-1B on the IOI and GP tasks, respectively.

For \textbf{IOI}, the circuit retains attention primarily in late layers
(8-11, 14-15), while MLP blocks are active across most of the network, except
for layers~5-6 and~12-13, which are fully pruned. This pattern mirrors the
MLP-dominant computation observed in GPT-2 for the same task, scaled to a
deeper architecture.

For \textbf{GP}, the circuit is substantially sparser. Attention activity is
distributed across layers 3-4, 6, 8-11, and 14, while MLP computation is
restricted to early layers (0-2), layer~10-11, 13, and~15. Multiple layers
are fully inactive (5, 7, 12), consistent with the localized processing
pattern observed in GPT-2 GP circuits.

\subsection{Llama 3.1-8B}

Tables~\ref{tab:pruning_summary_llama_8b_ioi} and~\ref{tab:pruning_summary_llama_8b_gp} present results for
Llama~3.1-8B. The 8B model supports stronger pruning overall: for GP, 75\% of
transformer blocks and 95.7\% of attention heads are pruned, representing the
most extreme sparsity observed in any of our experiments. This suggests that
the GP behavior in Llama~3.1-8B is supported by a remarkably small subnetwork
relative to the full model capacity.

For IOI on 8B, the circuit exhibits a characteristic split: attention is
concentrated in late layers (14-17, 23-27, 30-31), while MLPs are active
across most of the depth. This attention-sparse, MLP-dense pattern is
consistent across model scales and supports the conclusion that IOI relies
heavily on nonlinear residual-stream transformations throughout the network.

\begin{table*}[ht]
\centering
\small
\renewcommand{\arraystretch}{1.2}
\begin{tabular}{c|c|c|c|c|c|c}
\toprule
\textbf{Layer} & \textbf{Attn Block} & \textbf{MLP Block} & \textbf{Attn Heads} & \textbf{Attn Neurons} & \textbf{MLP Hidden} & \textbf{MLP Output} \\
\midrule
0  & Pruned & Active & 0/32  & 0/2048   & 4600/8192 & 992/2048  \\
1  & Pruned & Active & 0/32  & 0/2048   & 1962/8192 & 330/2048  \\
2  & Active & Active & 2/32  & 43/2048  & 1626/8192 & 489/2048  \\
3  & Pruned & Active & 0/32  & 0/2048   & 1510/8192 & 432/2048  \\
4  & Pruned & Active & 0/32  & 0/2048   & 1130/8192 & 299/2048  \\
5  & Pruned & Pruned & 0/32  & 0/2048   & 0/8192    & 0/2048    \\
6  & Pruned & Pruned & 0/32  & 0/2048   & 0/8192    & 0/2048    \\
7  & Pruned & Active & 0/32  & 0/2048   & 1405/8192 & 470/2048  \\
8  & Active & Active & 4/32  & 161/2048 & 1851/8192 & 827/2048  \\
9  & Active & Active & 7/32  & 281/2048 & 1973/8192 & 842/2048  \\
10 & Active & Active & 6/32  & 234/2048 & 2154/8192 & 753/2048  \\
11 & Active & Active & 11/32 & 403/2048 & 1608/8192 & 488/2048  \\
12 & Pruned & Pruned & 0/32  & 0/2048   & 0/8192    & 0/2048    \\
13 & Pruned & Pruned & 0/32  & 0/2048   & 0/8192    & 0/2048    \\
14 & Active & Active & 14/32 & 717/2048 & 1422/8192 & 736/2048  \\
15 & Active & Active & 14/32 & 661/2048 & 1952/8192 & 1042/2048          \\
\bottomrule
\end{tabular}
\caption{Node Pruning summary: attention blocks, MLP blocks, and retained components. Llama3.2-1B. IOI Taks.}
\label{tab:pruning_summary_llama_1b_ioi}
\end{table*}

\begin{table*}[ht]
\centering
\small
\renewcommand{\arraystretch}{1.2}
\begin{tabular}{c|c|c|c|c|c|c}
\toprule
\textbf{Layer} & \textbf{Attn Block} & \textbf{MLP Block} & \textbf{Attn Heads} & \textbf{Attn Neurons} & \textbf{MLP Hidden} & \textbf{MLP Output} \\
\midrule
0  & Pruned & Active & 0/32  & 0/4096   & 8283/14336  & 3005/4096 \\
1  & Active & Active & 4/32  & 293/4096 & 6206/14336  & 1489/4096 \\
2  & Active & Active & 3/32  & 132/4096 & 6360/14336  & 2082/4096 \\
3  & Pruned & Active & 0/32  & 0/4096   & 5169/14336  & 1752/4096 \\
4  & Pruned & Active & 0/32  & 0/4096   & 4236/14336  & 1752/4096 \\
5  & Pruned & Active & 0/32  & 0/4096   & 3643/14336  & 1594/4096 \\
6  & Pruned & Active & 0/32  & 0/4096   & 3226/14336  & 1547/4096 \\
7  & Pruned & Active & 0/32  & 0/4096   & 2934/14336  & 1334/4096 \\
8  & Pruned & Pruned & 0/32  & 0/4096   & 0/14336     & 0/4096    \\
9  & Pruned & Active & 0/32  & 0/4096   & 2367/14336  & 1175/4096 \\
10 & Active & Pruned & 3/32  & 240/4096 & 0/14336     & 0/4096    \\
11 & Pruned & Active & 0/32  & 0/4096   & 2409/14336  & 1195/4096 \\
12 & Pruned & Active & 0/32  & 0/4096   & 2817/14336  & 1549/4096 \\
13 & Pruned & Active & 0/32  & 0/4096   & 3244/14336  & 2210/4096 \\
14 & Active & Active & 6/32  & 531/4096 & 3807/14336  & 2679/4096 \\
15 & Active & Active & 7/32  & 507/4096 & 3953/14336  & 2776/4096 \\
16 & Active & Active & 7/32  & 697/4096 & 4249/14336  & 2827/4096 \\
17 & Active & Active & 6/32  & 480/4096 & 4415/14336  & 2850/4096 \\
18 & Pruned & Active & 0/32  & 0/4096   & 4325/14336  & 2110/4096 \\
19 & Pruned & Active & 0/32  & 0/4096   & 3985/14336  & 1822/4096 \\
20 & Pruned & Active & 0/32  & 0/4096   & 4055/14336  & 1731/4096 \\
21 & Pruned & Active & 0/32  & 0/4096   & 3868/14336  & 1768/4096 \\
22 & Pruned & Pruned & 0/32  & 0/4096   & 0/14336     & 0/4096    \\
23 & Active & Pruned & 10/32 & 731/4096 & 0/14336     & 0/4096    \\
24 & Active & Pruned & 10/32 & 863/4096 & 0/14336     & 0/4096    \\
25 & Active & Pruned & 8/32  & 822/4096 & 0/14336     & 0/4096    \\
26 & Active & Pruned & 8/32  & 606/4096 & 0/14336     & 0/4096    \\
27 & Active & Pruned & 6/32  & 690/4096 & 0/14336     & 0/4096    \\
28 & Pruned & Pruned & 0/32  & 0/4096   & 0/14336     & 0/4096    \\
29 & Pruned & Active & 0/32  & 0/4096   & 2511/14336  & 1497/4096 \\
30 & Active & Active & 9/32  & 957/4096 & 3403/14336  & 1968/4096 \\
31 & Active & Active & 5/32  & 451/4096 & 4338/14336  & 3095/4096 \\
\bottomrule
\end{tabular}
\caption{Node Pruning summary: attention blocks, MLP blocks, and retained components. Llama3.1-8B. IOI Task.}
\label{tab:pruning_summary_llama_8b_ioi}
\end{table*}

\begin{table*}[ht]
\centering
\small
\renewcommand{\arraystretch}{1.2}
\begin{tabular}{c|c|c|c|c|c|c}
\toprule
\textbf{Layer} & \textbf{Attn Block} & \textbf{MLP Block} & \textbf{Attn Heads} & \textbf{Attn Neurons} & \textbf{MLP Hidden} & \textbf{MLP Output} \\
\midrule
0  & Pruned & Active & 0/32  & 0/2048   & 4811/8192 & 1220/2048 \\
1  & Pruned & Active & 0/32  & 0/2048   & 4114/8192 & 977/2048  \\
2  & Pruned & Active & 0/32  & 0/2048   & 3604/8192 & 949/2048  \\
3  & Active & Pruned & 5/32  & 226/2048 & 0/8192    & 0/2048    \\
4  & Active & Pruned & 6/32  & 254/2048 & 0/8192    & 0/2048    \\
5  & Pruned & Pruned & 0/32  & 0/2048   & 0/8192    & 0/2048    \\
6  & Active & Pruned & 5/32  & 242/2048 & 0/8192    & 0/2048    \\
7  & Pruned & Pruned & 0/32  & 0/2048   & 0/8192    & 0/2048    \\
8  & Active & Active & 6/32  & 242/2048 & 2492/8192 & 758/2048  \\
9  & Active & Active & 8/32  & 327/2048 & 2795/8192 & 869/2048  \\
10 & Pruned & Active & 0/32  & 0/2048   & 2296/8192 & 753/2048  \\
11 & Active & Active & 2/32  & 116/2048 & 1843/8192 & 771/2048  \\
12 & Pruned & Pruned & 0/32  & 0/2048   & 0/8192    & 0/2048    \\
13 & Pruned & Active & 0/32  & 0/2048   & 1143/8192 & 684/2048  \\
14 & Pruned & Active & 0/32  & 0/2048   & 961/8192  & 705/2048  \\
15 & Pruned & Active & 0/32  & 0/2048   & 1261/8192 & 842/2048  \\
\bottomrule
\end{tabular}
\caption{Node Pruning summary: attention blocks, MLP blocks, and retained components. Llama3.2-1B. Gender Pronouns Task.}
\label{tab:pruning_summary_llama_1b_gp}
\end{table*}

\begin{table*}[ht]
\centering
\small
\renewcommand{\arraystretch}{1.2}
\begin{tabular}{c|c|c|c|c|c|c}
\toprule
\textbf{Layer} & \textbf{Attn Block} & \textbf{MLP Block} & \textbf{Attn Heads} & \textbf{Attn Neurons} & \textbf{MLP Hidden} & \textbf{MLP Output} \\
\midrule
0  & Pruned & Active & 0/32  & 0/4096   & 7765/14336  & 2862/4096 \\
1  & Pruned & Active & 0/32  & 0/4096   & 9052/14336  & 2552/4096 \\
2  & Pruned & Active & 0/32  & 0/4096   & 8827/14336  & 2825/4096 \\
3  & Active & Active & 6/32  & 560/4096 & 8455/14336  & 2452/4096 \\
4  & Active & Active & 8/32  & 765/4096 & 8144/14336  & 2622/4096 \\
5  & Pruned & Active & 0/32  & 0/4096   & 6984/14336  & 2308/4096 \\
6  & Active & Active & 6/32  & 649/4096 & 5914/14336  & 1709/4096 \\
7  & Active & Pruned & 8/32  & 630/4096 & 0/14336     & 0/4096    \\
8  & Pruned & Pruned & 0/32  & 0/4096   & 0/14336     & 0/4096    \\
9  & Pruned & Pruned & 0/32  & 0/4096   & 0/14336     & 0/4096    \\
10 & Pruned & Pruned & 0/32  & 0/4096   & 0/14336     & 0/4096    \\
11 & Pruned & Pruned & 0/32  & 0/4096   & 0/14336     & 0/4096    \\
12 & Pruned & Pruned & 0/32  & 0/4096   & 0/14336     & 0/4096    \\
13 & Active & Active & 10/32 & 841/4096 & 4396/14336  & 1697/4096 \\
14 & Pruned & Active & 0/32  & 0/4096   & 4486/14336  & 1773/4096 \\
15 & Pruned & Active & 0/32  & 0/4096   & 4369/14336  & 1718/4096 \\
16 & Active & Active & 9/32  & 714/4096 & 3776/14336  & 1577/4096 \\
17 & Pruned & Active & 0/32  & 0/4096   & 3645/14336  & 1823/4096 \\
18 & Active & Pruned & 6/32  & 481/4096 & 0/14336     & 0/4096    \\
19 & Pruned & Pruned & 0/32  & 0/4096   & 0/14336     & 0/4096    \\
20 & Pruned & Active & 0/32  & 0/4096   & 2355/14336  & 1598/4096 \\
21 & Pruned & Active & 0/32  & 0/4096   & 1926/14336  & 1368/4096 \\
22 & Pruned & Pruned & 0/32  & 0/4096   & 0/14336     & 0/4096    \\
23 & Active & Active & 3/32  & 238/4096 & 1416/14336  & 1452/4096 \\
24 & Pruned & Active & 0/32  & 0/4096   & 1050/14336  & 1079/4096 \\
25 & Pruned & Pruned & 0/32  & 0/4096   & 0/14336     & 0/4096    \\
26 & Pruned & Pruned & 0/32  & 0/4096   & 0/14336     & 0/4096    \\
27 & Pruned & Pruned & 0/32  & 0/4096   & 0/14336     & 0/4096    \\
28 & Pruned & Pruned & 0/32  & 0/4096   & 0/14336     & 0/4096    \\
29 & Pruned & Active & 0/32  & 0/4096   & 1092/14336  & 1028/4096 \\
30 & Pruned & Active & 0/32  & 0/4096   & 1632/14336  & 1655/4096 \\
31 & Pruned & Active & 0/32  & 0/4096   & 2444/14336  & 1906/4096 \\
\bottomrule
\end{tabular}
\caption{Node Pruning summary: attention blocks, MLP blocks, and retained components. Llama3.1-8B. Gender Pronouns Task}
\label{tab:pruning_summary_llama_8b_gp}
\end{table*}



\section{GPT2-XL}

\label{sec:gpt2-xl}
Tables~\ref{tab:pruning_summary_ioi_xl}-\ref{tab:pruning_summary_gt_xl} present per-layer node pruning
summaries for GPT-2-XL (1.5B parameters, 48 layers, 25 heads per layer) across
all three tasks. Table~\ref{tab:metrics_gpt_xl} reports the corresponding fidelity
metrics.

\paragraph{IOI.}
The GPT-2-XL IOI circuit is broadly distributed, with active attention in 28 of
48 layers and active MLPs in 36 of 48 layers. Despite this breadth, substantial
within-layer sparsity is achieved: on average, only 7.5 of 25 heads are
retained per active attention layer, and MLP hidden neurons are pruned by
approximately 62\%. Six layers are fully pruned (6, 10--15, 29, 34). The overall pattern confirms that IOI requires distributed computation across many layers, even in larger models, but with high intra-layer sparsity.

\paragraph{GP.}
GP on GPT-2-XL reveals an extreme sparsity pattern. Only 19 of 48 attention
layers retain any heads, and only 21 layers retain MLP neurons. A large
contiguous block of mid-layers (6-12, 16, 28-29, 34, 46-47) is fully
pruned. The retained attention concentrates in layers 13--27 with very few
heads per layer (1-6), while MLP activity shifts to late layers (30-45).
This suggests a two-phase computation: sparse attention-based feature
extraction in middle layers, followed by MLP-based decision-making in late
layers.

\paragraph{GT.}
GT produces the most extreme pruning on GPT-2-XL. All attention is
concentrated in exactly four layers (23, 25, 26, 28), each retaining all 25
heads, which suggests that GT computation is highly localized. All remaining layers have zero active attention heads. MLP computation is split
between early layers (0-5, 8-9) and late layers (30-45), with a large
fully-pruned gap (layers~6-7, 10--29 except for the four attention layers).
This extreme localization is consistent with the GPT-2 small GT circuit and
provides further evidence that numerical reasoning in transformers relies on a
small number of specialized layers.

\paragraph{Fidelity.}
All three tasks achieve strong fidelity on GPT-2-XL (Table~\ref{tab:metrics_gpt_xl}):
accuracy exceeds 95\% for all tasks, and KL divergence remains below 0.56.
The GT circuit achieves particularly low KL (0.0047), indicating near-perfect
distributional agreement despite extreme sparsity.

.

\begin{table*}[ht]
\centering
\small

\renewcommand{\arraystretch}{1.2}
\begin{tabular}{c|c|c|c|c|c|c}
\toprule
\textbf{Layer} & \textbf{Attn Block} & \textbf{MLP Block} & \textbf{Attn Heads} & \textbf{Attn Neurons} & \textbf{MLP Hidden} & \textbf{MLP Output} \\
\midrule
0  & Active & Active & 2/25  & 59/1600  & 5851/6400 & 1600/1600 \\
1  & Pruned & Active & 0/25  & 0/1600   & 3232/6400 & 1555/1600 \\
2  & Active & Active & 6/25  & 254/1600 & 3051/6400 & 1461/1600 \\
3  & Active & Active & 7/25  & 299/1600 & 2905/6400 & 1388/1600 \\
4  & Active & Active & 7/25  & 318/1600 & 2789/6400 & 1198/1600 \\
5  & Pruned & Active & 0/25  & 0/1600   & 2649/6400 & 1064/1600 \\
6  & Pruned & Pruned & 0/25  & 0/1600   & 0/6400    & 0/1600    \\
7  & Pruned & Active & 0/25  & 0/1600   & 2459/6400 & 867/1600  \\
8  & Active & Pruned & 6/25  & 293/1600 & 0/6400    & 0/1600    \\
9  & Pruned & Active & 0/25  & 0/1600   & 2353/6400 & 760/1600  \\
10 & Pruned & Pruned & 0/25  & 0/1600   & 0/6400    & 0/1600    \\
11 & Pruned & Pruned & 0/25  & 0/1600   & 0/6400    & 0/1600    \\
12 & Pruned & Pruned & 0/25  & 0/1600   & 0/6400    & 0/1600    \\
13 & Pruned & Pruned & 0/25  & 0/1600   & 0/6400    & 0/1600    \\
14 & Active & Pruned & 4/25  & 210/1600 & 0/6400    & 0/1600    \\
15 & Pruned & Pruned & 0/25  & 0/1600   & 0/6400    & 0/1600    \\
16 & Pruned & Active & 0/25  & 0/1600   & 2326/6400 & 778/1600  \\
17 & Active & Pruned & 9/25  & 359/1600 & 0/6400    & 0/1600    \\
18 & Active & Active & 10/25 & 404/1600 & 2341/6400 & 871/1600  \\
19 & Pruned & Active & 0/25  & 0/1600   & 2520/6400 & 985/1600  \\
20 & Active & Active & 6/25  & 231/1600 & 2542/6400 & 1055/1600 \\
21 & Pruned & Active & 0/25  & 0/1600   & 2574/6400 & 1045/1600 \\
22 & Active & Active & 8/25  & 324/1600 & 2377/6400 & 959/1600  \\
23 & Active & Active & 8/25  & 318/1600 & 2473/6400 & 967/1600  \\
24 & Active & Active & 7/25  & 265/1600 & 2403/6400 & 922/1600  \\
25 & Active & Active & 11/25 & 439/1600 & 2527/6400 & 957/1600  \\
26 & Active & Active & 10/25 & 376/1600 & 2383/6400 & 908/1600  \\
27 & Active & Pruned & 8/25  & 292/1600 & 0/6400    & 0/1600    \\
28 & Active & Active & 10/25 & 372/1600 & 2347/6400 & 844/1600  \\
29 & Pruned & Pruned & 0/25  & 0/1600   & 0/6400    & 0/1600    \\
30 & Active & Active & 9/25  & 334/1600 & 2083/6400 & 743/1600  \\
31 & Active & Active & 10/25 & 422/1600 & 2063/6400 & 768/1600  \\
32 & Active & Pruned & 10/25 & 376/1600 & 0/6400    & 0/1600    \\
33 & Active & Pruned & 14/25 & 527/1600 & 0/6400    & 0/1600    \\
34 & Pruned & Pruned & 0/25  & 0/1600   & 0/6400    & 0/1600    \\
35 & Active & Active & 10/25 & 363/1600 & 1889/6400 & 783/1600  \\
36 & Active & Active & 10/25 & 384/1600 & 1976/6400 & 804/1600  \\
37 & Active & Active & 9/25  & 316/1600 & 1963/6400 & 812/1600  \\
38 & Active & Active & 13/25 & 483/1600 & 1878/6400 & 779/1600  \\
39 & Active & Active & 15/25 & 576/1600 & 1903/6400 & 765/1600  \\
40 & Pruned & Active & 0/25  & 0/1600   & 2019/6400 & 785/1600  \\
41 & Pruned & Active & 0/25  & 0/1600   & 2020/6400 & 778/1600  \\
42 & Active & Active & 12/25 & 423/1600 & 2157/6400 & 809/1600  \\
43 & Active & Active & 10/25 & 334/1600 & 2104/6400 & 802/1600  \\
44 & Active & Active & 10/25 & 340/1600 & 2238/6400 & 833/1600  \\
45 & Active & Active & 9/25  & 290/1600 & 2331/6400 & 806/1600  \\
46 & Pruned & Active & 0/25  & 0/1600   & 2429/6400 & 846/1600  \\
47 & Pruned & Active & 0/25  & 0/1600   & 2551/6400 & 886/1600  \\
\bottomrule
\end{tabular}
\caption{Node Pruning(Ours) summary for Indirect Object Identification, GPT-2-XL Model: attention blocks, MLP blocks, and retained components.}
\label{tab:pruning_summary_ioi_xl}
\end{table*}

\begin{table*}[ht]
\centering
\small
\renewcommand{\arraystretch}{1.2}
\begin{tabular}{c|c|c|c|c|c|c}
\toprule
\textbf{Layer} & \textbf{Attn Block} & \textbf{MLP Block} & \textbf{Attn Heads} & \textbf{Attn Neurons} & \textbf{MLP Hidden} & \textbf{MLP Output} \\
\midrule
0 & Pruned & Active & 0/25 & 0/1600 & 4058/6400 & 1358/1600 \\
1 & Active & Active & 5/25 & 234/1600 & 1669/6400 & 708/1600 \\
2 & Pruned & Active & 0/25 & 0/1600 & 1460/6400 & 513/1600 \\
3 & Pruned & Active & 0/25 & 0/1600 & 1337/6400 & 404/1600 \\
4 & Pruned & Active & 0/25 & 0/1600 & 1341/6400 & 348/1600 \\
5 & Pruned & Active & 0/25 & 0/1600 & 1173/6400 & 264/1600 \\
6 & Pruned & Pruned & 0/25 & 0/1600 & 0/6400 & 0/1600 \\
7 & Pruned & Pruned & 0/25 & 0/1600 & 0/6400 & 0/1600 \\
8 & Pruned & Pruned & 0/25 & 0/1600 & 0/6400 & 0/1600 \\
9 & Active & Pruned & 1/25 & 48/1600 & 0/6400 & 0/1600 \\
10 & Pruned & Pruned & 0/25 & 0/1600 & 0/6400 & 0/1600 \\
11 & Pruned & Pruned & 0/25 & 0/1600 & 0/6400 & 0/1600 \\
12 & Pruned & Pruned & 0/25 & 0/1600 & 0/6400 & 0/1600 \\
13 & Active & Pruned & 2/25 & 89/1600 & 0/6400 & 0/1600 \\
14 & Active & Pruned & 4/25 & 178/1600 & 0/6400 & 0/1600 \\
15 & Active & Pruned & 4/25 & 183/1600 & 0/6400 & 0/1600 \\
16 & Pruned & Pruned & 0/25 & 0/1600 & 0/6400 & 0/1600 \\
17 & Active & Pruned & 3/25 & 144/1600 & 0/6400 & 0/1600 \\
18 & Active & Pruned & 6/25 & 277/1600 & 0/6400 & 0/1600 \\
19 & Active & Pruned & 6/25 & 263/1600 & 0/6400 & 0/1600 \\
20 & Active & Pruned & 4/25 & 174/1600 & 0/6400 & 0/1600 \\
21 & Active & Pruned & 4/25 & 182/1600 & 0/6400 & 0/1600 \\
22 & Active & Active & 6/25 & 288/1600 & 818/6400 & 409/1600 \\
23 & Active & Pruned & 3/25 & 131/1600 & 0/6400 & 0/1600 \\
24 & Active & Pruned & 5/25 & 219/1600 & 0/6400 & 0/1600 \\
25 & Active & Pruned & 4/25 & 180/1600 & 0/6400 & 0/1600 \\
26 & Active & Pruned & 3/25 & 138/1600 & 0/6400 & 0/1600 \\
27 & Active & Pruned & 1/25 & 35/1600 & 0/6400 & 0/1600 \\
28 & Pruned & Pruned & 0/25 & 0/1600 & 0/6400 & 0/1600 \\
29 & Pruned & Pruned & 0/25 & 0/1600 & 0/6400 & 0/1600 \\
30 & Active & Active & 1/25 & 36/1600 & 827/6400 & 547/1600 \\
31 & Pruned & Active & 0/25 & 0/1600 & 775/6400 & 467/1600 \\
32 & Pruned & Active & 0/25 & 0/1600 & 698/6400 & 341/1600 \\
33 & Pruned & Active & 0/25 & 0/1600 & 627/6400 & 490/1600 \\
34 & Pruned & Pruned & 0/25 & 0/1600 & 0/6400 & 0/1600 \\
35 & Pruned & Active & 0/25 & 0/1600 & 684/6400 & 477/1600 \\
36 & Active & Active & 1/25 & 29/1600 & 591/6400 & 530/1600 \\
37 & Pruned & Active & 0/25 & 0/1600 & 548/6400 & 322/1600 \\
38 & Pruned & Active & 0/25 & 0/1600 & 531/6400 & 410/1600 \\
39 & Pruned & Active & 0/25 & 0/1600 & 572/6400 & 552/1600 \\
40 & Pruned & Active & 0/25 & 0/1600 & 414/6400 & 434/1600 \\
41 & Pruned & Active & 0/25 & 0/1600 & 529/6400 & 357/1600 \\
42 & Pruned & Active & 0/25 & 0/1600 & 514/6400 & 302/1600 \\
43 & Pruned & Active & 0/25 & 0/1600 & 537/6400 & 468/1600 \\
44 & Active & Pruned & 1/25 & 24/1600 & 0/6400 & 0/1600 \\
45 & Pruned & Active & 0/25 & 0/1600 & 510/6400 & 308/1600 \\
46 & Pruned & Pruned & 0/25 & 0/1600 & 0/6400 & 0/1600 \\
47 & Pruned & Pruned & 0/25 & 0/1600 & 0/6400 & 0/1600 \\
\bottomrule
\end{tabular}
\caption{Node Pruning(Ours) summary for Gender Pronouns, GPT-2-XL Model: attention blocks, MLP blocks, and retained components.}
\label{tab:pruning_summary_gp_xl}
\end{table*}

\begin{table*}[ht]
\centering
\small
\renewcommand{\arraystretch}{1.2}
\begin{tabular}{c|c|c|c|c|c|c}
\toprule
\textbf{Layer} & \textbf{Attn Block} & \textbf{MLP Block} & \textbf{Attn Heads} & \textbf{Attn Neurons} & \textbf{MLP Hidden} & \textbf{MLP Output} \\
\midrule
0 & Pruned & Active & 0/25 & 0/1600 & 4372/6400 & 1598/1600 \\
1 & Pruned & Active & 0/25 & 0/1600 & 2737/6400 & 1578/1600 \\
2 & Pruned & Active & 0/25 & 0/1600 & 3306/6400 & 1557/1600 \\
3 & Pruned & Active & 0/25 & 0/1600 & 3270/6400 & 1549/1600 \\
4 & Pruned & Active & 0/25 & 0/1600 & 3111/6400 & 1538/1600 \\
5 & Pruned & Active & 0/25 & 0/1600 & 3179/6400 & 1541/1600 \\
6 & Pruned & Pruned & 0/25 & 0/1600 & 0/6400 & 0/1600 \\
7 & Pruned & Pruned & 0/25 & 0/1600 & 0/6400 & 0/1600 \\
8 & Pruned & Active & 0/25 & 0/1600 & 2770/6400 & 1518/1600 \\
9 & Pruned & Active & 0/25 & 0/1600 & 2782/6400 & 1522/1600 \\
10 & Pruned & Pruned & 0/25 & 0/1600 & 0/6400 & 0/1600 \\
11 & Pruned & Pruned & 0/25 & 0/1600 & 0/6400 & 0/1600 \\
12 & Pruned & Pruned & 0/25 & 0/1600 & 0/6400 & 0/1600 \\
13 & Pruned & Pruned & 0/25 & 0/1600 & 0/6400 & 0/1600 \\
14 & Pruned & Pruned & 0/25 & 0/1600 & 0/6400 & 0/1600 \\
15 & Pruned & Pruned & 0/25 & 0/1600 & 0/6400 & 0/1600 \\
16 & Pruned & Pruned & 0/25 & 0/1600 & 0/6400 & 0/1600 \\
17 & Pruned & Pruned & 0/25 & 0/1600 & 0/6400 & 0/1600 \\
18 & Pruned & Pruned & 0/25 & 0/1600 & 0/6400 & 0/1600 \\
19 & Pruned & Pruned & 0/25 & 0/1600 & 0/6400 & 0/1600 \\
20 & Pruned & Pruned & 0/25 & 0/1600 & 0/6400 & 0/1600 \\
21 & Pruned & Pruned & 0/25 & 0/1600 & 0/6400 & 0/1600 \\
22 & Pruned & Pruned & 0/25 & 0/1600 & 0/6400 & 0/1600 \\
23 & Active & Pruned & 25/25 & 1559/1600 & 0/6400 & 0/1600 \\
24 & Pruned & Pruned & 0/25 & 0/1600 & 0/6400 & 0/1600 \\
25 & Active & Pruned & 25/25 & 1562/1600 & 0/6400 & 0/1600 \\
26 & Active & Pruned & 25/25 & 1557/1600 & 0/6400 & 0/1600 \\
27 & Pruned & Pruned & 0/25 & 0/1600 & 0/6400 & 0/1600 \\
28 & Active & Pruned & 25/25 & 1553/1600 & 0/6400 & 0/1600 \\
29 & Pruned & Pruned & 0/25 & 0/1600 & 0/6400 & 0/1600 \\
30 & Pruned & Active & 0/25 & 0/1600 & 2774/6400 & 1557/1600 \\
31 & Pruned & Active & 0/25 & 0/1600 & 2789/6400 & 1565/1600 \\
32 & Pruned & Active & 0/25 & 0/1600 & 3003/6400 & 1554/1600 \\
33 & Pruned & Active & 0/25 & 0/1600 & 3016/6400 & 1579/1600 \\
34 & Pruned & Active & 0/25 & 0/1600 & 3030/6400 & 1555/1600 \\
35 & Pruned & Active & 0/25 & 0/1600 & 3222/6400 & 1570/1600 \\
36 & Pruned & Active & 0/25 & 0/1600 & 3287/6400 & 1567/1600 \\
37 & Pruned & Active & 0/25 & 0/1600 & 3321/6400 & 1558/1600 \\
38 & Pruned & Active & 0/25 & 0/1600 & 3349/6400 & 1571/1600 \\
39 & Pruned & Active & 0/25 & 0/1600 & 3330/6400 & 1563/1600 \\
40 & Pruned & Active & 0/25 & 0/1600 & 3069/6400 & 1549/1600 \\
41 & Pruned & Active & 0/25 & 0/1600 & 3024/6400 & 1565/1600 \\
42 & Pruned & Active & 0/25 & 0/1600 & 2952/6400 & 1554/1600 \\
43 & Pruned & Active & 0/25 & 0/1600 & 2714/6400 & 1533/1600 \\
44 & Pruned & Active & 0/25 & 0/1600 & 2446/6400 & 1519/1600 \\
45 & Pruned & Active & 0/25 & 0/1600 & 2336/6400 & 1495/1600 \\
46 & Pruned & Pruned & 0/25 & 0/1600 & 0/6400 & 0/1600 \\
47 & Pruned & Pruned & 0/25 & 0/1600 & 0/6400 & 0/1600 \\
\bottomrule
\end{tabular}
\caption{Node Pruning(Ours) summary for Greater Than, GPT-2-XL Model: attention blocks, MLP blocks, and retained components.}
\label{tab:pruning_summary_gt_xl}
\end{table*}

\begin{table}[ht]
\centering
\begin{tabular}{lccc}
\hline
\textbf{Metric} & \textbf{GT} & \textbf{GP} & \textbf{IOI} \\
\hline
Accuracy & 0.9501 & 0.9798 & 0.9752 \\
KL Divergence / Faithfulness & 0.0047 & 0.2336 & 0.5568 \\
Logit/Prob Diff & 0.4562 & 4.5936 & 4.2321 \\
\hline
\end{tabular}
\caption{Fidelity metrics across GT, GP, and IOI. GPT2-XL model}
\label{tab:metrics_gpt_xl}
\end{table}